\pdfoutput=1

\documentclass[11pt]{article}

\usepackage[preprint]{acl}

\usepackage{times}
\usepackage{latexsym}

\usepackage[T1]{fontenc}

\usepackage[utf8]{inputenc}

\usepackage{microtype}

\usepackage{inconsolata}

\usepackage{graphicx}

\usepackage{amsmath}
\usepackage{amssymb}
\usepackage{array}
\usepackage{colortbl}
\usepackage{hhline}
\usepackage{mathtools}
\usepackage{subcaption}
\usepackage{xcolor}
\usepackage{booktabs}
\definecolor{lightergreen}{RGB}{26, 153, 0}
\definecolor{lighteryellow}{RGB}{230, 132, 34}
\definecolor{lighterred}{RGB}{204, 0, 0}
\definecolor{high_green}{RGB}{1, 54, 27}
\definecolor{mid_green}{RGB}{44, 156, 22}
\definecolor{low_orange}{RGB}{222, 114, 4}

\title{Investigating the Transferability of Code Repair \\ for Low-Resource Programming Languages}

\author{
 \textbf{Kyle Wong \textsuperscript{1}},
 \textbf{Alfonso Amayuelas \textsuperscript{1}},
 \textbf{Liangming Pan \textsuperscript{2}},
 \textbf{William Yang Wang \textsuperscript{1}}
\\
\textsuperscript{1} University of California, Santa Barbara \\
\textsuperscript{2}University of Arizona \\
\texttt{\{knw, amayuelas\}@ucsb.edu} \\
\texttt{liangmingpan@arizona.edu, william@cs.ucsb.edu}
}

\begin{document}
\maketitle

\begin{abstract}
Large language models (LLMs) have shown remarkable performance on code generation tasks. A recent use case is iterative code repair, where an LLM fixes an incorrect program by rationalizing about errors and generating new code. Recent works augment the code repair process by integrating modern techniques such as chain-of-thought reasoning or distillation, but only study their benefits on high-resource languages like Python, and ignore low-resource languages like Perl. To address this gap of knowledge, we investigate the benefits of distilling code repair for both high and low resource languages to determine if the techniques that are effective in a high resource setting are also applicable in a low resource setting. Our evaluation shows that distilling the ability to repair code has language dependent benefits. To explain this behavior, we perform a further analysis and find that contrary to preexisting beliefs, the correlation between reasoning ability and code correction ability is weak. We hypothesize this weak correlation is magnified in low-resource settings where base models lack deep knowledge of a programming language, leading to wavering benefits of code repair.
\end{abstract}

\section{Introduction}

\newcommand{\unfootnote}[1]{%
  \begingroup
  \renewcommand{\thefootnote}{\relax}%
  \footnotetext{#1}%
  \endgroup
}

\unfootnote{Source code: \href{https://github.com/KyleWong288/Distill_LRPL}{github.com/KyleWong288/Distill\_LRPL}}

\begin{figure*}[ht]
  \centering
  \includegraphics[width=\textwidth]{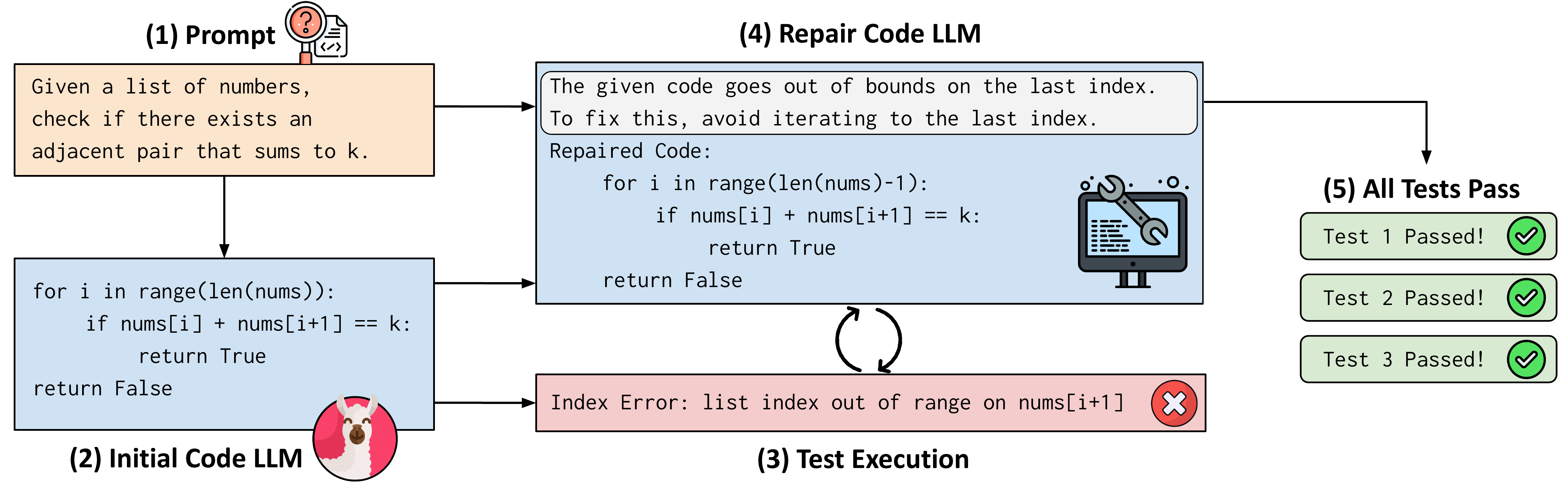}
  \caption{A standard code repair framework. In (1) and (2), a code LLM is given a question and generates a solution. In (3),  test cases are executed and an error message is extracted. In (4), a repair LLM is given the question, incorrect solution, and error message, and generates a repair. A repair contains a rationale explaining why the old code was incorrect and how to fix it, followed by new code. If the new code is still incorrect, we iteratively generate new repairs using the code from previous repairs. In (5), we stop when all tests pass or after a fixed number of iterations.}
  \label{fig:repair_pipeline}

\end{figure*}

While large language models (LLMs) like GPT-4 \citep{openai2024gpt4} display remarkable coding abilities on popular benchmarks like HumanEval \citep{chen2021humaneval}, the performance of smaller models like CodeLlama-7b \citep{rozière2024code} lag behind. Thus, frameworks that improve code generation and can apply to smaller models have become increasingly useful (\citealp{ding2024cycle}; \citealp{shinn2023reflexion}). Code repair is one such framework, which is inspired by the editing process of human programmers: erroneous feedback is provided through executing tests, while programmers rationalize about those errors to fix the code. The standard code repair pipeline we adopt is depicted in Figure \ref{fig:repair_pipeline}.

Although code repair is seemingly effective, recent works conclude that it is bottlenecked by a model's underlying ability to rationalize about errors \citep{olausson2024selfrepair}, leading to lesser improvements on weaker models. To further improve repairs for smaller LLMs, prior works transfer the ability to reason about incorrect code, either from humans \citep{chen2024ilf} or other LLMs (\citealp{olausson2024selfrepair}; \citealp{ren2024reflectioncoder}). However, the effects of knowledge transfer on code repair is primarily studied on high-resource programming languages (HRPLs), and its generalizability to low-resource programming languages (LRPLs) remains under-explored.

Distilling code repair is particularly useful in a low-resource setting because it improves code accuracy without requiring more pretraining data, which is the main limitation for LRPLs. For example, a modern code LLM DeepSeek-Coder \citep{guo2024deepseekcoder} is trained on a dataset scraped from Github repositories, containing high-resource languages like Python and Java at rates of 15.12\% an 18.63\%, while other languages like Perl and Golang are at much lesser rates of 0.1\% and 0.32\%. Since code repair is effective at increasing pass rates without extra pretraining data, understanding when it works best is critical for improving low-resource code generation.

Our work investigates if the code repair techniques that work on HRPLs are equally applicable to LRPLs. If transferring reasoning abilities can sufficiently enhance LRPL code repair, then it could mitigate the problem of low representation in pretraining data. Thus, we aim to understand the efficacy of distilling code repair for LRPLs and offer a guideline on whether to prioritize reasoning or code knowledge. To do so, we compare two approaches to distillation: the first is transferring only reasoning (rationale-only), while the second is transferring both reasoning and code completions (rationale-plus-code). 

We hypothesize that even if repair models are given a high quality rationale, they still often fail to fix incorrect code because they lack knowledge on the syntax and semantics of a programming language. This weakness is magnified in low-resource settings, so it is more beneficial to transfer code completions for LRPLs, but less necessary for HRPLs. To verify that the benefits of distilling code repair are language dependent, we perform a comprehensive suite of experiments and present our main research questions and findings below.

\begin{itemize}

\item \textbf{How effective is distillation for LRPLs?} Distilled repair models lead to higher pass rates. We see a relative increase in the average pass@1 of CodeLlama-7b-Instruct by 99.5\% for Perl, 112.8\% for Golang, and 144.5\% for Swift after four rounds of repair on HumanEval. 

\item \textbf{How effective is distillation for LRPLs compared to HRPLs?} Rationale-plus-code distillation outperforms rationale-only distillation on LRPLs, but provides negligible improvements on HRPLs. On HumanEval, we see a relative increase in the pass@1 by 21.9\% for Perl, 11.0\% for Golang, and 16.3\% for Swift, but see smaller increases on HRPLs. 

\item \textbf{Why does transferring both reasoning and code only outperform on LRPLs?}
Even if repair models have a good rationale, they still struggle to make accurate code changes. This weakness is magnified in LRPLs, where models have a weaker understanding of the language. On LRPLs, rationale-only repair models generate a correct rationale 91.1\% of the time, but only generate correct code 10.0\% of the time, exposing a weak correlation between rationale correctness and code correctness.

\end{itemize}

\section{Related Work}


\subsection{Repairing Code with LLMs}
\label{sec:repair_framework_list}
Iterative repair through feedback has been a well-studied area, surveyed in \citep{pan2023automatically} and \citep{fernandes2023bridging}. The scenarios where self correction works best is also surveyed in \citep{kamoi2024self_correct_survey}. For code repair in specific, frameworks like Self-Repair \citep{olausson2024selfrepair}, CYCLE \citep{ding2024cycle}, CodeChain \citep{le2024codechain}, ILF \citep{chen2024ilf}, Self-Edit \citep{zhang2023selfedit}, Self-Debugging \citep{chen2023teaching}, and Reflexion \citep{shinn2023reflexion} have shown promising increases in pass rates.

\subsection{Distillation for Code Repair}
\label{sec:related_works_distillation}
Distillation is the process of transferring knowledge from high capacity models to lower capacity models. Previous works show distillation can transfer the ability to reason and generate code (\citealp{sun2024enhancing}; \citealp{wei2023magicoder}; \citealp{xu2023wizardlm}; \citealp{luo2023wizardcoder}; \citealp{li2022explanations}), but transferring the ability to repair code remains less explored. Recent methods like PERsD \citep{chen2024personalised} and LLM2LLM \citep{lee2024llm2llm} use distillation to augment fine-tuning datasets. Self-Repair \citep{olausson2024selfrepair} also experiments with transferring rationales from GPT-4 to CodeLlama-13b-Instruct in-context to improve the repair process. However, none of these approaches investigates the efficacy of distilling repair for low-resource code.

\subsection{Low-Resource Programming Languages}
Code repair experiments are usually evaluated on high-resource languages like Python, but our work investigates its efficacy on different languages. For evaluation, many works (\citealp{athiwaratkun2023multilingual}; \citealp{orlanski2023babelcode}; \citealp{zheng2023codegeex}) have created datasets to benchmark code generation in a multilingual setting. Since finding human written low-resource code is difficult, other approaches use capable LLMs to synthetically create low-resource code. Works like MultiPL-T \citep{cassano2024multiPLT} and MultiPL-E \citep{cassano2022multiPLE} translate popular pre-training datasets and monolingual benchmarks into a wide variety of different programming languages. Other works also study the transferability of coding ability between different languages (\citealp{baltaji2024language_transfers}; \citealp{gong2022multicoder}).

\section{Methodology}

We provide an overview of a standard code repair framework, followed by an explanation of our distillation from a teacher model to a student model.

\subsection{Code Repair Framework}
\label{sec:code_repair_framework}
We adopt a standard code repair pipeline as the basis of our method. While there exist more complex frameworks like those referenced in Section \ref{sec:repair_framework_list}, our goal is to study the benefits of knowledge distillation, as opposed to inventing a new framework. Thus, we use a basic code repair pipeline as demonstrated in Figure \ref{fig:repair_pipeline}, where the main components are the initial code generation, test execution, and iterative repair. We provide a formal explanation for each component. 

First, we define $M_{init}$ as the model generating initial answers. For a question $q$, we obtain $n \geq 10$ initial samples, because it allows us to compute pass@10, along with lower variance pass@1 and pass@5 estimates. We define $c_{t,i}$ as the $i$-th code sample generated on repair round $t$, where $t=0$ denotes the initial generation. Obtaining the initial code generations is formalized in expression \ref{eq:initial_gen}.
\begin{equation}
\label{eq:initial_gen}
    M_{init}(q) \to \{c_{0,i}\}_{i=1}^{n}
\end{equation}

Next, we define $E$ as the code executor. Given a set of code samples, we execute the test cases associated with $q$ on each sample. This produces a set of error messages, where $e_{t,i}$ is the error message resulting from $c_{t,i}$. Obtaining the error messages is formalized in expression \ref{eq:error}.
\begin{equation}
\label{eq:error}
    E(q, \{c_{t,i}\}_{i=1}^{n}) \to \{e_{t,i}\}_{i=1}^{n}
\end{equation}

Finally, we define $M_{repair}$ as the model generating repairs. $M_{repair}$ has the same underlying model architecture as $M_{init}$. A repair is composed of a chain-of-thought \citep{wei2023chainofthought} rationale $r_{t,i}$, and the associated code completion $c_{t,i}$. Our work compares two different scenarios: transferring only reasoning (rationale-only distillation) vs transferring both reasoning and code completions (rationale-plus-code distillation).

For rationale-only distillation, we transfer reasoning through in-context learning \citep{brown2020few_shot} by generating the rationale $r_{t,i}$ from a separate larger model $M_{teacher}$. Obtaining a repair is formalized in expressions \ref{eq:rationale_teacher} and \ref{eq:repair_icl}. 
\begin{equation}
\label{eq:rationale_teacher}
    M_{teacher}(q, c_{t,i}, e_{t,i}) \to r_{t+1, i}
\end{equation}
\begin{equation}
\label{eq:repair_icl}
    M_{repair}(q, c_{t,i}, e_{t,i}, r_{t+1, i}) \to c_{t+1,i}
\end{equation}

For rationale-plus-code distillation, $M_{repair}$ is responsible for independently generating both the rationale and code completion, and obtaining a repair is formalized in expression \ref{eq:repair_incorrect}.
\begin{equation}
\label{eq:repair_incorrect}
    M_{repair}(q, c_{t,i}, e_{t,i}) \to (r_{t+1, i}, c_{t+1,i})
\end{equation}

\begin{figure*}[ht]
  \centering
  \includegraphics[width=\textwidth]{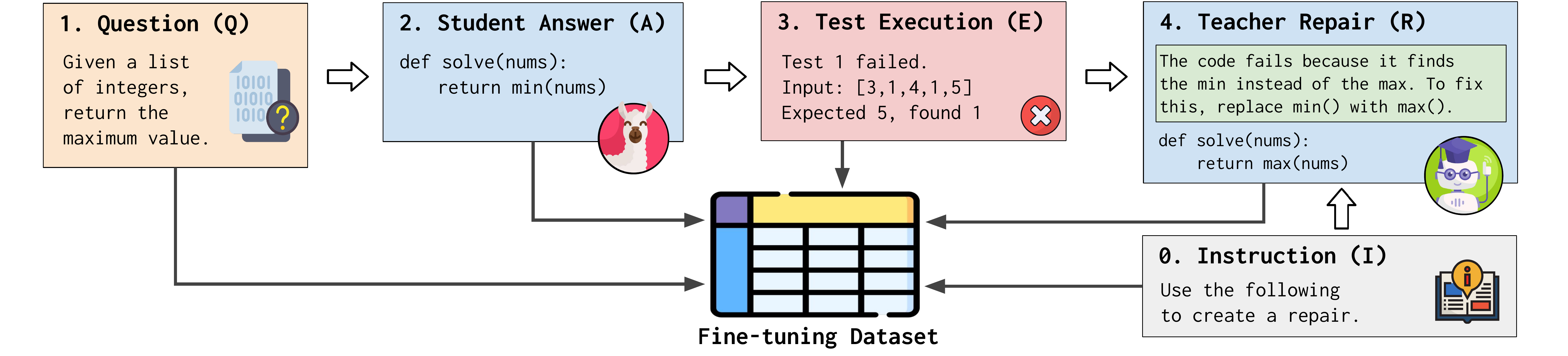}
  \caption{Our dataset construction pipeline. Examples in the fine-tuning dataset contain an instruction, the original question, the student's incorrect answer, the execution feedback, and the teacher's correct repair.}
  \label{fig:dataset_construction}
\end{figure*}

\subsection{Dataset Construction}
\label{sec:dataset_construction}

To perform rationale-plus-code distillation from a teacher model to a student model, we construct a fine-tuning dataset. Our teacher model is GPT-3.5-Turbo (\citealp{ouyang2022gpt_3.5}; \citealp{openai2022chatgpt}), while our student models are CodeLlama-7b-Instruct \citep{rozière2024code}, CodeLlama-7b \citep{rozière2024code}, and Mistral-7b \citep{jiang2023mistral}. The fine-tuning datasets are constructed from MBXP \citep{athiwaratkun2023multilingual}, which consists of multiple language specific benchmarks, each containing around 960 questions with corresponding test cases. An artificial train-test split is created by taking 800 random examples as potential training data and reserving the rest for testing. We process potential training examples into a finalized dataset, visualized in Figure \ref{fig:dataset_construction}. Our dataset is formally composed of five-tuples in the form $(I, Q, A, E, R)$, which we further explain.

\paragraph{Instruction and Question.} 
Each five-tuple begins with a constant instruction $I$, informing the model to perform code repair. Next is a question $Q$, containing a problem description and function declaration. We collect $Q$ by directly using the prompts provided in MBXP.

\paragraph{Answer and Error.}
The student's incorrect answer is represented with $A$, which is collected by prompting a student model with $Q$. To ensure $A$ is incorrect, we allow the student to continually generate independent samples, which are then immediately tested. Once a sample fails the given test cases, we select that sample as $A$. Then, we collect the associated error message $E$ from the execution feedback.

\paragraph{Repair.}
Lastly, we finish with $R$, the teacher model's repair. We collect $R$ by providing a teacher model with $(I, Q, A, E)$ and prompting it to generate two main components. The first component is a rationale that explains why the error occurred and a plan to fix it. The second component is updated code based on $A$, denoted with $A'$. To ensure $A'$ is correct, we allow the teacher to continually generate independent repairs, which are then immediately tested. Once $A'$ passes the given test cases, we select the associated repair as $R$. Our prompt format can be examined in Appendix \ref{app:prompt_repair}

\paragraph{Quantity of Examples.}
Although the original train split starts with 800 examples, our construction pipeline results in fine-tuning datasets with around 400 examples. Referencing Figure \ref{fig:dataset_construction}, this is because we may fail to obtain a usable $A$ in step (2) or a usable $R$ in step (4). In step (2), student models may consistently generate correct code. We allow a maximum of 10 samples before discarding the current example. Conversely, in step (4), teacher models may consistently generate incorrect code. We allow a maximum of 20 samples before discarding the current example. When prompting the teacher model, we use few-shot prompting \citep{brown2020few_shot} with three examples as an attempt to generate better repairs. The exact dataset sizes are listed in Appendix \ref{app:ft_dataset_sizes}.

\section{Experiment}

Our goal is to understand and compare the transferability of code repair for HRPLs and LRPLs, so we conduct a comprehensive experiment with three high-resource and low-resource languages. We identify Python, Javascript, and Java as high-resource, and identify Perl, Golang, and Swift as low-resource. These languages are picked based on having the highest three and lowest three pass rates observed in the original MBXP evaluations \citep{athiwaratkun2023multilingual}, as well as cross referencing the percentage of each language in DeepSeek-Coder's pretraining dataset \citep{guo2024deepseekcoder}, since it loosely reflects the distribution of programming languages found on Github. For each language, we perform our dataset construction and fine-tune a student model. Then, we generate an initial round of output and perform four rounds of code repair.

\subsection{Experimental Setup}
\label{sec:experimental_setup}

\paragraph{Models.} 
To show our observations generalize to non-instruction-tuned and different model families, we run our experiments on CodeLlama-7b-Instruct \citep{rozière2024code}, CodeLlama-7b \citep{rozière2024code}, and Mistral-7b \citep{jiang2023mistral}. These models are used for the initial generation, and then a fine-tuned version of the same architecture is used as the distilled repair model.

\paragraph{Benchmarks.}
We evaluate on our MBXP \citep{athiwaratkun2023multilingual} test split from Section \ref{sec:dataset_construction}, containing around 160 programming problems for each language. Additionally, we evaluate on MultiLingual HumanEval \citep{athiwaratkun2023multilingual}, a variation of HumanEval \citep{chen2021humaneval} transcompiled to different languages, which also contains around 160 programming problems for each language. Our evaluation on MultiLingual HumanEval (HumanEval for brevity) shows that fine-tuned repair models generalize to other datasets.

\paragraph{Metrics.}
We evaluate all generations using pass@k \citep{chen2021humaneval}, a standard performance metric for code generation tasks. Since pass@k is prone to high variance, we use the unbiased estimator for pass@k, which estimates the probability that at least one out of k samples is correct. Given $n \geq k$ code samples where $c$ are correct, we compute pass@k using Equation \ref{eq:pass@k}.
\begin{equation}
\label{eq:pass@k}
    \text{pass@}k \coloneqq \mathop{\mathbb{E}}\limits_{\text{Problems}}\left[1 - \frac{\binom{n-c}{k}}{\binom{n}{k}}\right]
\end{equation}

\paragraph{Training and Inference Details.}
We perform a 90/10 train-dev split on the dataset resulting from Section \ref{sec:dataset_construction}, and train via LoRA fine-tuning \citep{hu2021lora}. During the initial generation, we sample 10 answers for each question and compute pass@k using n=10, allowing us to measure certain baselines. However, we only perform code repair on the first 5 samples for later repair rounds and compute pass@1 using n=5, because we only care about the repair pass@1. To encourage diversity between samples, we use nucleus sampling with a threshold of 0.95 and sampling temperature of 0.2. Further training and inference hyperparameters are listed in Appendix \ref{app:hyperparams}. For baselines that use a non-fine-tuned model for repair, we use one-shot prompting, whose format is shown in Appendix \ref{app:prompt_repair}.

\subsection{Baselines}

\begin{figure*}[ht]
    \centering
    \begin{subfigure}[b]{\textwidth}
        \centering
        \includegraphics[width=0.85\textwidth]{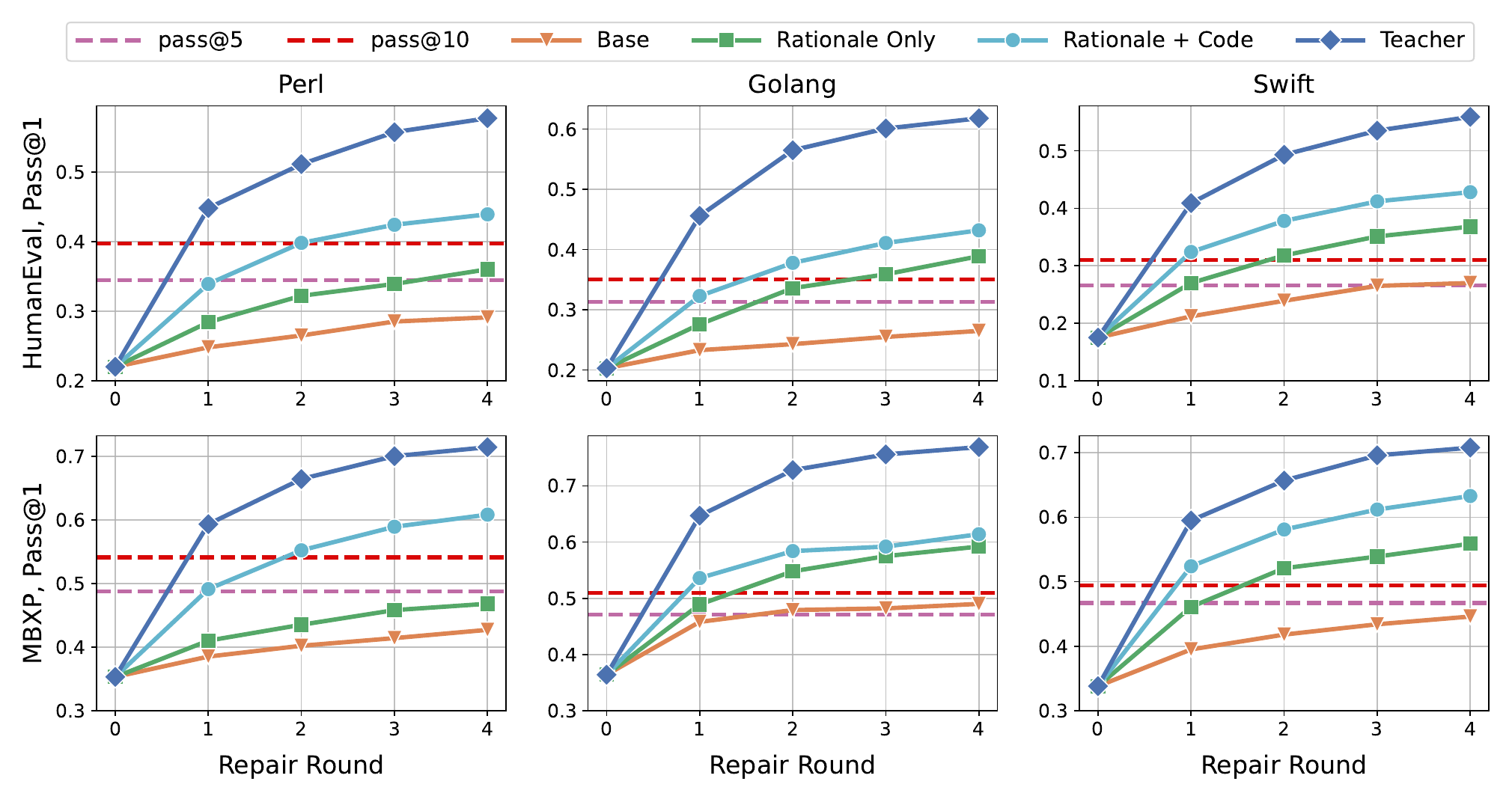}
        \caption{Low-resource languages}
    \end{subfigure}
    \begin{subfigure}[b]{\textwidth}
        \centering
        \includegraphics[width=0.85\textwidth]{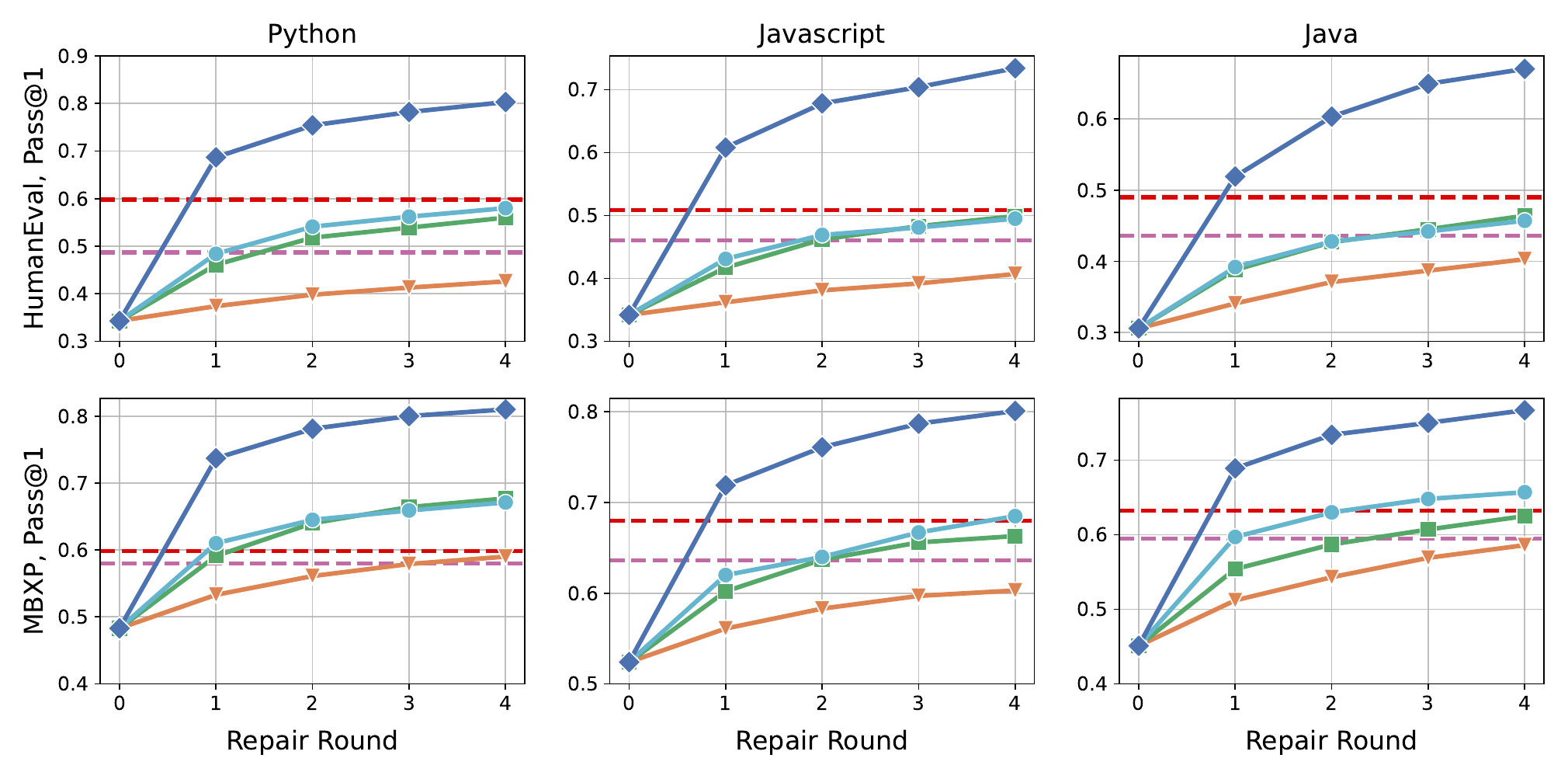}
        \caption{High-resource languages}
    \end{subfigure}
    \caption{Mean pass@1 versus repair round for CodeLlama-7b-Instruct. Round 0 denotes the initial generation. Rationale-plus-code distillation outperforms rationale-only distillation on low-resource languages, but performs similarly on high-resource languages.}
    \label{fig:plots_cli}
\end{figure*}

We compare the pass@1 of rationale-plus-code distillation to five different baselines. These baselines demonstrate how different repair approaches perform on HRPLs vs LRPLs, allowing us to analyze trade-offs and scenarios where rationale-plus-code distillation works best.

\paragraph{Non-repair Independent Sampling.}
We compare the efficiency of iterative repair rounds with independent sampling. This provides insights if distilled models can achieve higher pass rates with equal or fewer inference calls. Our experiment conducts 1 initial generation and 4 repair rounds for a total of 5 inference calls, so we compare the final pass@1 with the pass@5 and pass@10 of the initial generations.

\paragraph{Basic Iterative Repair.}
To measure the benefits of distillation, we evaluate how well the base model performs on iterative repair without any extra modifications. We use the same code repair workflow, but replace the distilled repair model with its non-fine-tuned counterpart.

\paragraph{Rationale-only distillation}
To measure the necessity of transferring code completions, we adopt an adjacent idea from Self-Repair \citep{olausson2024selfrepair}, where only the rationalization about an error is transferred in-context. First, a teacher model is prompted to generate the rationale portion of a repair. Then, a base repair model is prompted to generate the code portion of a repair, with the teacher's rationale appended in-context. We use the same teacher and student models as Section \ref{sec:experimental_setup}, and our prompt to extract the teacher's rationale is in Appendix \ref{app:prompt_icl}.

\paragraph{Teacher Repair.}
For demonstrating the limitations of distillation, we use the same code repair framework, but replace the student model with the teacher model used during dataset construction. This acts as a rough upper bound for the student model, and illustrates potential room for improvement.

\subsection{Results}
Our experiments provide empirical results demonstrating the pass@1 improvements of distilling both rationales and code completions, along with a wavering benefit of code repair between HRPLs and LRPLs. We report our results on CodeLlama-7b-Instruct in Figure \ref{fig:plots_cli}, and similar results on CodeLlama-7b and Mistral-7b can be found in Appendix \ref{app:evaluation_clb} and \ref{app:evaluation_mist}.

\paragraph{Impact of Distillation.}
\emph{Distilling code repair consistently outperforms repair with a base model.} Figure \ref{fig:plots_cli} shows both distillation plot lines steadily trending higher than the base repair plot lines. One explanation for this is that higher quality rationales may causally influence code correctness. Thus, weaker base models may not benefit as much from a framework like code repair, which requires strong reasoning to diagnose mistakes.

\paragraph{Distilled Repair vs Independent Sampling.}
\emph{Distilled repair achieves higher pass rates than independently sampling.} Across all languages, four rounds of distilled code repair outperforms the initial pass@5. Furthermore, rationale-plus-code distilled models considerably outperform the initial pass@10 on LRPLs shown in Figure \ref{fig:plots_cli}. Thus, when limited to a small amount of inference calls, distilling code repair can be a more efficient alternative than independent sampling for increasing pass rates.

\begin{table}[htbp]
  \centering
  \resizebox{\columnwidth}{!}{
  \begin{tabular}{l|ccc}
    \hline
    \multicolumn{4}{c}{\textbf{HumanEval Pass@1}} \\
    \hline
    \small Language & \small Initial & \small Rationale Only & \small Rationale + Code \\
    [-2.5pt]
             & \small Round & \small Distillation & \small Distillation \\
    \hline
    Perl         & 0.220       & 0.360     & \textbf{0.439} \small \textcolor{mid_green}{$\hspace{0.2em}\uparrow$21.9\%} \\ 
    Golang       & 0.203       & 0.389     & \textbf{0.432} \small \textcolor{mid_green}{$\hspace{0.2em}\uparrow$11.3\%} \\
    Swift        & 0.175       & 0.368     & \textbf{0.428} \small \textcolor{mid_green}{$\hspace{0.2em}\uparrow$16.3\%} \\
    \hline
    Python       & 0.343       & 0.560     & \textbf{0.580} \small \textcolor{mid_green}{$\hspace{0.2em}\uparrow$3.57\%} \\
    Javascript   & 0.342       & \textbf{0.499}   & 0.495 \small \textcolor{orange}{$\hspace{0.2em}\downarrow$0.80\%} \\
    Java         & 0.306       & \textbf{0.464} & 0.457 \small \textcolor{orange}{$\hspace{0.2em}\downarrow$1.50\%} \\
    \hline
    \multicolumn{4}{c}{\textbf{MBXP Pass@1}} \\
    \hline
    Perl         & 0.353       & 0.468     & \textbf{0.608} \small \textcolor{mid_green}{$\hspace{0.2em}\uparrow$29.9\%} \\
    Golang       & 0.364       & 0.592     & \textbf{0.614} \small \textcolor{mid_green}{$\hspace{0.2em}\uparrow$3.71\%} \\
    Swift        & 0.338       & 0.559     & \textbf{0.633} \small \textcolor{mid_green}{$\hspace{0.2em}\uparrow$13.2\%} \\
    \hline
    Python       & 0.483       & \textbf{0.677}  & 0.671 \small \textcolor{orange}{$\hspace{0.2em}\downarrow$0.88\%}\\
    Javascript   & 0.524       & 0.663  & \textbf{0.685} \small \textcolor{mid_green}{$\hspace{0.2em}\uparrow$3.31\%} \\
    Java         & 0.451       & 0.625  & \textbf{0.657} \small \textcolor{mid_green}{$\hspace{0.2em}\uparrow$5.12\%} \\
    \hline 
  \end{tabular}
  }
  \caption{Mean pass@1 of initial generations vs the two variants of distillation. Relative percent increases between the two variants of distillation is also provided. Rationale-plus-code distillation consistently outperforms on LRPLs, but performs around the same or slightly worse on HRPLs.}
  \label{tab:increase_cli}
\end{table}

\paragraph{Benefits on HRPLs vs LRPLs.}

\emph{Rationale-plus-code distillation consistently outperforms rationale-only distillation on LRPLs, but fails to do the same on HRPLs.} Although it might seem intuitive that transferring rationale-only would have lesser benefits than transferring rationale-plus-code, Table \ref{tab:increase_cli} quantifies how this is not always the case. Our results indicate distilling code repair provides a wavering benefit depending on the programming language, and spurs us to investigate why.

\section{Analysis}

After comparing the results of rationale-plus-code distillation with rationale-only distillation, we observe the benefits of distilling code repair depends on whether the language is high-resource or low-resource. Thus, we seek an explanation on why rationale-plus-code distillation achieves higher pass rates on LRPLs, but not HRPLs. Previous research (\citealp{olausson2024selfrepair}; \citealp{ren2024reflectioncoder}) hypothesizes that code repair is bottlenecked by the model's underlying ability to create a high quality rationale, which our experimental results support. However, there remains a lacking explanation of why repair models still generate incorrect code, even when given a high quality rationale.

We hypothesize there exists a second bottleneck: even if repair models are given good rationales, they fail to fix incorrect code because they lack the knowledge to convert suggestions from the rationale into accurate code modifications. This effect is magnified in a low-resource setting because base models are less knowledgeable about a language's syntax and semantics, explaining why transferring code completions leads to more potent benefits for LRPLs. To support our hypothesis, we analyze the quality of rationales and a repair model's knowledge of a language.


\newcolumntype{C}[1]{>{\centering\arraybackslash}p{#1}} 
\newcolumntype{L}[1]{>{\arraybackslash}p{#1}} 

\begin{table*}[]
    \centering
    \begin{tabular}{c}
        \hline
        \begin{subtable}[t]{\textwidth}
            \centering
            \begin{tabular}{C{3cm} || C{5.77cm} ||  C{6cm}} 
            & \textsc{Rationale + Code} & \textsc{Rationale Only} \\
            \end{tabular}
            \begin{tabular}{C{3cm} || C{2cm}|C{2cm}|C{0.9cm} || C{2cm}|C{2cm}|C{0.9cm}}
                \hline
                 & \textbf{Code Fails} & \textbf{Code Passes} & \textbf{Total} & \textbf{Code Fails} & \textbf{Code Passes} & \textbf{Total} \\
                 \hline
                \multicolumn{7}{c}{\textsc{lrpls}} \\
                \hline
                \textbf{Bad Rationale} & 12.4\% & 1.0\% & 13.4\% & 8.4\% & 0.5\% & 8.9\% \\
                \hline
                \textbf{Good Rationale} & 71.2\% & \textbf{15.4\%} & 86.6\% & 81.6\% & \textbf{9.5\%} & 91.1\%\\
                \hline
                \textbf{Total} & 83.6\% & 16.4\% & & 90.0\% & 10.0\% &  \\
                \hline
            \end{tabular}
        \end{subtable}
        \\
        \begin{subtable}[t]{\textwidth}
            \centering
            \begin{tabular}{C{3cm} ||C{2cm}|C{2cm}|C{0.9cm} || C{2cm}|C{2cm}|C{0.9cm}}
                \multicolumn{7}{c}{\textsc{hrpls}} \\
                \hline
                \textbf{Bad Rationale} & 19.7\% & 2.3\% & 22.0\% & 9.3\% & 0.7\% & 10.0\% \\
                \hline
                \textbf{Good Rationale} & 63.9\% & \textbf{14.1\%} & 78.0\% & 75.9\% & \textbf{14.1\%} & 90.0\% \\
                \hline
                \textbf{Total} & 83.6\% & 16.4\% &  & 85.2\% & 14.8\% & \\
                \hline
            \end{tabular}
        \end{subtable}
    \end{tabular}
    
    \caption{Empirical relationship between rationale quality and code correctness. Even when a good rationale is provided, the rate of producing passing code is significantly less than the rate of producing failing code. This exposes a weak correlation between rationale quality and code correctness.}
    \label{tab:rationale_vs_code}
\end{table*}

\subsection{Correlation between Rationale and Code}
To support our hypothesis that a bottleneck exists in a model's ability to convert reasoning into code, we analyze the relationship between rationale quality and code correctness in Table \ref{tab:rationale_vs_code}. We quantitatively show that repair models frequently generate correct rationales, yet still generate incorrect code, exposing a weak correlation between the reasoning process and code editing process.

We use GPT-4 as an LLM judge to determine whether a rationale is sufficient or insufficient. Although human evaluation would be preferred, crowdsourcing participants well-versed in programming languages like Perl and Swift and capable of solving the coding problems found in HumanEval is challenging. We picked GPT-4 because the distilled rationales were generated with GPT-3.5-Turbo, and we aimed to use a more advanced and reliable model for better assessments \citep{zheng2023llmjudge}.

To obtain judgements, we present a question, incorrect code, error message, and rationale to GPT-4, and instruct it to produce a verdict. A rationale is labelled good if it contains accurate information and includes sufficient detail to repair the given code, and bad otherwise. Our judgement prompt can be found in Appendix \ref{app:prompt_judgement}. We obtain a verdict for all HumanEval rationales extracted between the initial generation and the first repair round.

Table \ref{tab:rationale_vs_code} demonstrates that the rate of a good rationale leading to passing code is consistently low. For LRPLs, rationale-only models have a rate of 9.5\%, while rationale-plus-code models have a higher rate of 15.4\%. However, for HRPLs, both distilled variants have equal rates of 14.1\%. Thus, these results expose how transferring code completions mitigates the weak correlation between rationale and code more effectively on LRPLs than for HRPLs.

Many code repair frameworks follow some variation of the same steps: obtain an error, rationalize, generate code. Although prior works (\citealp{chen2024ilf}; \citealp{olausson2024selfrepair}) focus solely on improving the weak link between error and rationale, our results highlight there is another weak link between rationale and code correctness. Since models have poor ability in converting a correct rationale into correct code modifications, augmenting the reasoning process alone leads to limited benefits. We display various examples where repair models provide good rationales, but the resulting code has clear syntactic or semantic errors in Appendix \ref{app:actual_examples}. 

\begin{table}[t]
  \centering
  \resizebox{\columnwidth}{!}{
  \begin{tabular}{l | clll}
    \hline
    \multicolumn{5}{c}{\textbf{HumanEval Average Syntax Errors}} \\
    \hline
    \multicolumn{1}{c|}{} & \multicolumn{1}{c}{Initial} & \multicolumn{1}{c}{Base} & \multicolumn{1}{c}{Rationale} & \multicolumn{1}{c}{Rationale} \\
    \multicolumn{1}{c|}{Language} & \multicolumn{1}{c}{Errors} & \multicolumn{1}{c}{Repair} & \multicolumn{1}{c}{Only} & \multicolumn{1}{c}{+ Code} \\
    \hline
    Perl         & 14.5 & 15.4 \small \textcolor{lighterred}{$\uparrow$0.9} & 17.8 \small \textcolor{lighterred}{$\uparrow$3.3} & \textbf{9.20} \small \textcolor{lightergreen}{$\downarrow$5.3} \\
    
    Golang       & 44.7 & 70.4 \small \textcolor{lighterred}{$\uparrow$25.7} & 48.7 \small \textcolor{lighterred}{$\uparrow$4.0} & \textbf{26.6} \small \textcolor{lightergreen}{$\downarrow$18.1} \\
    Swift        & 81.0  & 58.0 \small \textcolor{lightergreen}{$\downarrow$23.0} & 50.4 \small \textcolor{lightergreen}{$\downarrow$30.6} & \textbf{37.4} \small \textcolor{lightergreen}{$\downarrow$43.6} \\
    \hline
    Python       & 12.1 & 15.6 \small \textcolor{lighterred}{$\uparrow$3.5} & 18.2 \small \textcolor{lighterred}{$\uparrow$6.1} & \textbf{14.2} \small \textcolor{lighterred}{$\uparrow$2.1} \\
    Javascript   & 9.10 & 9.80 \small \textcolor{lighterred}{$\uparrow$0.7} & 27.6 \small \textcolor{lighterred}{$\uparrow$18.5} & \textbf{9.00} \small \textcolor{lightergreen}{$\downarrow$0.1} \\
    Java         & 39.6 & 41.2 \small \textcolor{lighterred}{$\uparrow$1.6} & \textbf{37.0} \small \textcolor{lightergreen}{$\downarrow$2.6} & 41.2 \small \textcolor{lighterred}{$\uparrow$1.6} \\
    \hline
    \multicolumn{5}{c}{\textbf{MBXP Average Syntax Errors}} \\
    \hline
    Perl         & 12.1 & 9.50 \small \textcolor{lightergreen}{$\downarrow$2.6} & 13.7 \small \textcolor{lighterred}{$\uparrow$1.6} & \textbf{2.70} \small \textcolor{lightergreen}{$\downarrow$9.4} \\
    Golang       & 33.2 & 29.2 \small \textcolor{lightergreen}{$\downarrow$4.0} & 26.8 \small \textcolor{lightergreen}{$\downarrow$6.4} & \textbf{14.6} \small \textcolor{lightergreen}{$\downarrow$18.6} \\
    Swift        & 60.4 & 36.0 \small \textcolor{lightergreen}{$\downarrow$24.0} & 27.8 \small \textcolor{lightergreen}{$\downarrow$32.6} & \textbf{11.0} \small \textcolor{lightergreen}{$\downarrow$49.4} \\
    \hline
    Python       & 1.80 & 5.20 \small \textcolor{lighterred}{$\uparrow$3.4} & 5.10 \small \textcolor{lighterred}{$\uparrow$3.3} & \textbf{3.60} \small \textcolor{lighterred}{$\uparrow$1.8} \\
    Javascript   & 4.60 & 4.20 \small \textcolor{lightergreen}{$\downarrow$0.4} & 11.8 \small \textcolor{lighterred}{$\uparrow$7.6} & \textbf{3.60} \small \textcolor{lightergreen}{$\downarrow$1.0} \\
    Java         & 29.2 & 26.4 \small \textcolor{lightergreen}{$\downarrow$2.8} & 21.4 \small \textcolor{lightergreen}{$\downarrow$5.0} & \textbf{20.4} \small \textcolor{lightergreen}{$\downarrow$8.8} \\
    \hline
  \end{tabular}
  }
  \caption{Mean number of syntax errors after code repair, along with the differences between the initial and final amount of errors. Rationale-plus-code distillation has a notably higher decline in syntax errors on LRPLs, but performs closer to baselines on HRPLs.}
  \label{tab:syntax_errors}
\end{table}

\subsection{Knowledge of LRPLs}
Lastly, we analyze why transferring code completions only achieves consistent improvements on LRPLs. To support our hypothesis that a base model's weak responsiveness is magnified in a low-resource setting, Table \ref{tab:syntax_errors} shows how rationale-plus-code distilled models exhibit a deeper understanding of LRPLs. We use the frequency of syntax errors as a proxy for knowledge, since generating code with syntax errors is a blatant sign that a model lacks comprehension of a language.

To measure this, we first extract the set of syntax errors from a particular code repair run. Syntax errors are those occurring before execution and caught during compilation or interpretation time. We can conveniently filter out syntax errors by parsing the execution feedback. Next, we compute the average amount of errors within the final repair round. Note that non-syntax errors can transform into syntax errors when repair models attempt to update code, leading to occasional increases. The average number of syntax errors for CodeLlama-7b-Instruct can be seen in Table \ref{tab:syntax_errors}, and similar results on CodeLlama-7b and Mistral-7b can be seen in Appendix \ref{tab:syntax_clb} and \ref{tab:syntax_mist}. 

For LRPLs, the decrease in syntax errors when transferring code completions is higher than the other baselines. Averaging over the 3 LRPLs, rationale-plus-code models finish with 16.9 errors, while rationale-only models finish with 30.9 errors. Thus, for the case of LRPLs, boosting rationale quality alone is not enough for encouraging a base model to generate a correct repair, and distilling code completions improves a model's knowledge on a programming language.

For HRPLs, transferring code completions leads to less potent improvements. The decrease in syntax errors is smaller, and the final amount of errors among all baselines are relatively close. Averaging over the 3 HRPLs, rationale-plus-code models finish with 15.3 errors, while rationale-only models finish with 20.2 errors. Furthermore, even the base repair model performs comparably, finishing with an average of 17.1 errors. Thus, for the case of HRPLs, transferring code completions is less necessary because the base model already has sufficient knowledge on a programming language.

\section{Conclusion}
We distilled the ability to repair code and demonstrated that transferring only rationales is sufficient for high-resource languages, but further transferring code completions is necessary for low-resource languages. We also exposed that the correlation between rationale quality and code correctness is lower than previously perceived, especially in low-resource settings. Rationale-plus-code distillation mitigates this weakness by improving a model's understanding of a programming language, resulting in better responsiveness to feedback. Further research in distillation is important because it allows smaller models to gain fluency without costly human labeling, creating efficient and high-performing LLMs suitable for consumer-grade devices. Such advancements would democratize the benefits of closed source research, making better code generation accessible for a wider range of languages, applications, and users.

\section*{Limitations}

One limitation is the lack of more challenging multilingual datasets. Other popular coding benchmarks like APPS \citep{hendrycks2021apps} and CodeContests \citep{Li_2022_codecontests} provide harder problems, which may demand stronger reasoning, but are only available in high-resource languages. Studying the benefits of distilling rationales on more reasoning heavy questions in low-resource languages would be an insightful future evaluation.

A natural limitation is the lack of instruction tuning datasets for LRPLs, as our fine-tuning datasets only contain around 400 examples. Although it would be ideal to have more examples, the amount of data available for these low-resource languages is naturally low. Our work counters concerns about the generalizability of our findings by evaluating on multiple models, languages, and benchmarks.

Another limitation in our evaluation are the stochastic processes within training and inference. To the best of our ability, we mitigate variance in our evaluation by seeding our training and inference, and by using the unbiased estimator of pass@k. However, since we use nucleus sampling for decoding, we observe there can be slight variations in our results. 

Lastly, an underlying limitation is our hardware for training and inference. We use Nvidia Titan RTX GPUs with 24GB memory, so the size of student models that we can fine-tune is limited, which is why we choose 7b models for our experiments. Furthermore, since our evaluation has many dimensions (6 languages, 3 models, 5 baselines, 2 benchmarks, 160 questions each benchmark), we are limited in the amount of sampling we can do for each question. Although it may be interesting to obtain higher pass@k rates like k=10 or k=100, these are not time efficient to measure and do not contribute that much to our arguments. Thus, we choose to only show pass@1 for repair rounds.


\section*{Ethics Statement}
Since computing resources and research funding is extremely valuable, querying costly models like GPT-4 should be conducted responsibly. Estimating costs before running experiments and making necessary adjustments is a responsible and resource-conscious approach to using such APIs. 

Furthermore, there exists the possibility that users apply code repair for harmful applications. People with malicious intentions could use our research to improve code generation in certain domains that produce dangerous code, such as attacks on privacy and security. We encourage that code repair be used for socially responsible technology.

\bibliography{custom}

\appendix
\onecolumn
\label{sec:appendix}

\section{Fine-tuning Dataset Sizes}
\label{app:ft_dataset_sizes}

\begin{table}[ht]
  \centering
  \resizebox{0.5\textwidth}{!}{
  \begin{tabular}{l | ccc | ll}
    \hline
    \multicolumn{6}{c}{\textbf{Fine-tuning Dataset Sizes}} \\
    \hline
    Language & Initial & Post-Student & Post-Teacher & Train & Dev \\
    \hline
    \multicolumn{6}{c}{\textbf{CodeLlama-7b-Instruct}} \\
    \hline
    Perl        & 800 & 649	& 489 & 440 & 49 \\
    Golang      & 800 & 601	& 455 & 409 & 46 \\
    Swift       & 800 & 635	& 470 & 423 & 47 \\
    \hline
    Python        & 800 & 559	& 446 &	401 & 45 \\
    Javascript    & 800 & 509	& 394 & 354 & 40 \\
    Java          & 800 & 667	& 510 &	459 & 51 \\
    \hline
    \multicolumn{6}{c}{\textbf{CodeLlama-7b}} \\
    \hline
    Perl        & 800 & 680	& 489 & 440 & 49 \\
    Golang      & 800 & 614	& 456 & 410 & 46 \\
    Swift       & 800 & 651	& 465 & 418 & 47 \\
    \hline
    Python      & 800 & 596	& 470 & 423 & 47 \\
    Javascript  & 800 & 586	& 470 & 423 & 47 \\
    Java        & 800 & 642	& 499 & 449 & 50 \\
    \hline
    \multicolumn{6}{c}{\textbf{Mistral-7b}} \\
    \hline
    Perl        & 800 & 689	& 533 & 479 & 54 \\
    Golang      & 800 & 745	& 539 & 459 & 54 \\
    Swift       & 800 & 625	& 468 & 421 & 47 \\
    \hline
    Python      & 800 & 602	& 487 & 438 & 49 \\
    Javascript  & 800 & 535	& 413 & 371 & 42 \\
    Java        & 800 & 573	& 439 & 395 & 44 \\
    \hline
  \end{tabular}
  }
  \caption{The final fine-tuning dataset sizes for each model, starting from the original MBXP train split of 800 questions. Intermediate sizes at each step of our dataset construction are also provided.}
  \label{tab:dataset_sizes}
\end{table}

\section{Training and Inference Hyperparameters}
\label{app:hyperparams}

We provide our training and inference hyperparameters used throughout experiments. All training and inference are conducted on Nvidia Titan RTX (24GB) GPUs.

For training, we use LoRA fine-tuning with a rank of 128, lora alpha of 128, lora dropout of 0.1, maximum sequence length of 2048, batch size of 4, gradient accumulation steps of 2, weight decay of 0.01, cosine learning rate scheduler with warm up steps of 10, and checkpoint every 50 steps. For models in the CodeLlama family, we train for 8 epochs with a learning rate of 2e-5, and for Mistral-7b, we train for 5 epochs with a learning rate of 5e-6. To obtain our final distilled repair model, we pick the checkpoint with the lowest validation loss. 

For inference, we use nucleus sampling with a threshold of 0.95, sampling temperature of 0.2, and limit the maximum new tokens to 800. When generating the initial 10 samples, we use random seeds from 1 to 10. For later repair rounds, all generations use a random seed of 17.

\section{In-Context Rationale Prompt}
\label{app:prompt_icl}
\begin{figure}[hb]
    \centering
    \includegraphics[width=0.8\textwidth]{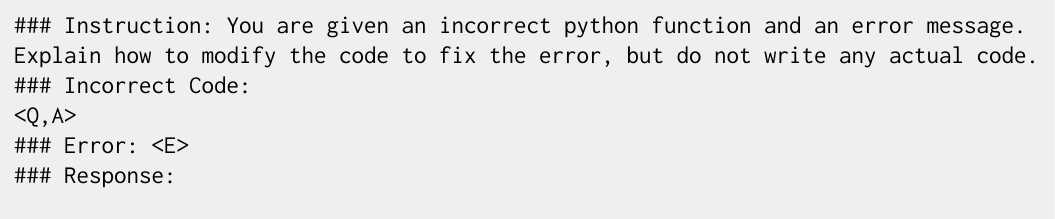}
    \caption{The prompt given to GPT-3.5-Turbo to generate the rationale portion of a repair. This is only used for the in-context learning baseline. <Q,A> is replaced with the question and previous answer, while <E> is replaced with the corresponding error.}
\end{figure}

\newpage
\section{Repair Prompt}
\label{app:prompt_repair}

We provide the general format of our repair prompt. When conducting code repair with a rationale-plus-code distilled model, we use zero-shot prompting. When conducting code repair with baselines, we use one-shot prompting. When creating a fine-tuning dataset with the teacher model, we use three-shot prompting. The few-shot examples change with each programming language.

\begin{figure}[hb]
    \centering
    \includegraphics[width=0.75\textwidth]{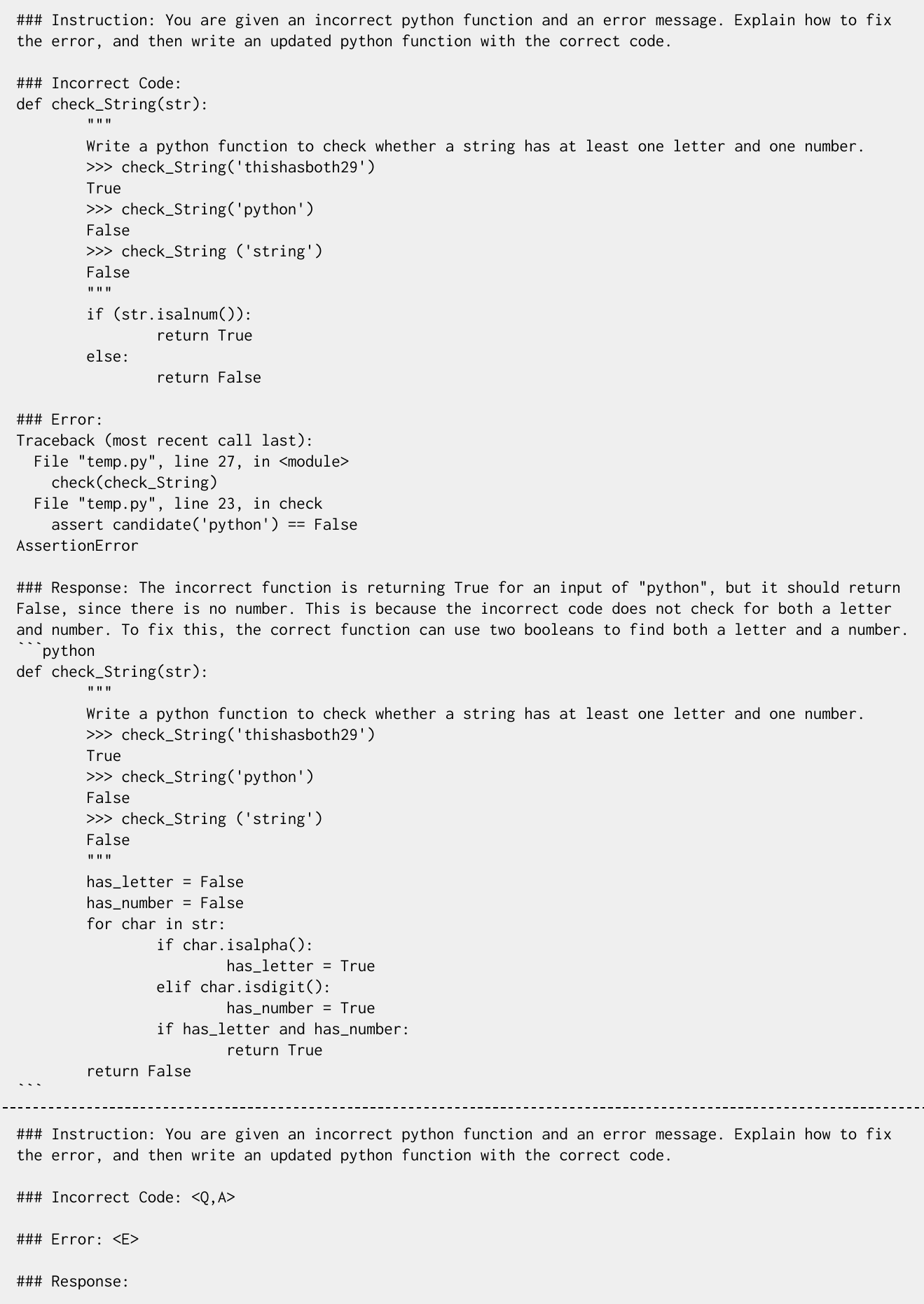}
    \caption{The prompt for generating a repair. For brevity, we only show a one-shot example. <Q,A> is replaced with the question and previous answer, while <E> is replaced with the corresponding error.}
\end{figure}

\newpage
\section{Evaluation on CodeLlama-7b}
\label{app:evaluation_clb}
\begin{figure}[htbp]
    \centering
    \begin{subfigure}[b]{\textwidth}
        \centering
        \includegraphics[width=0.95\textwidth]{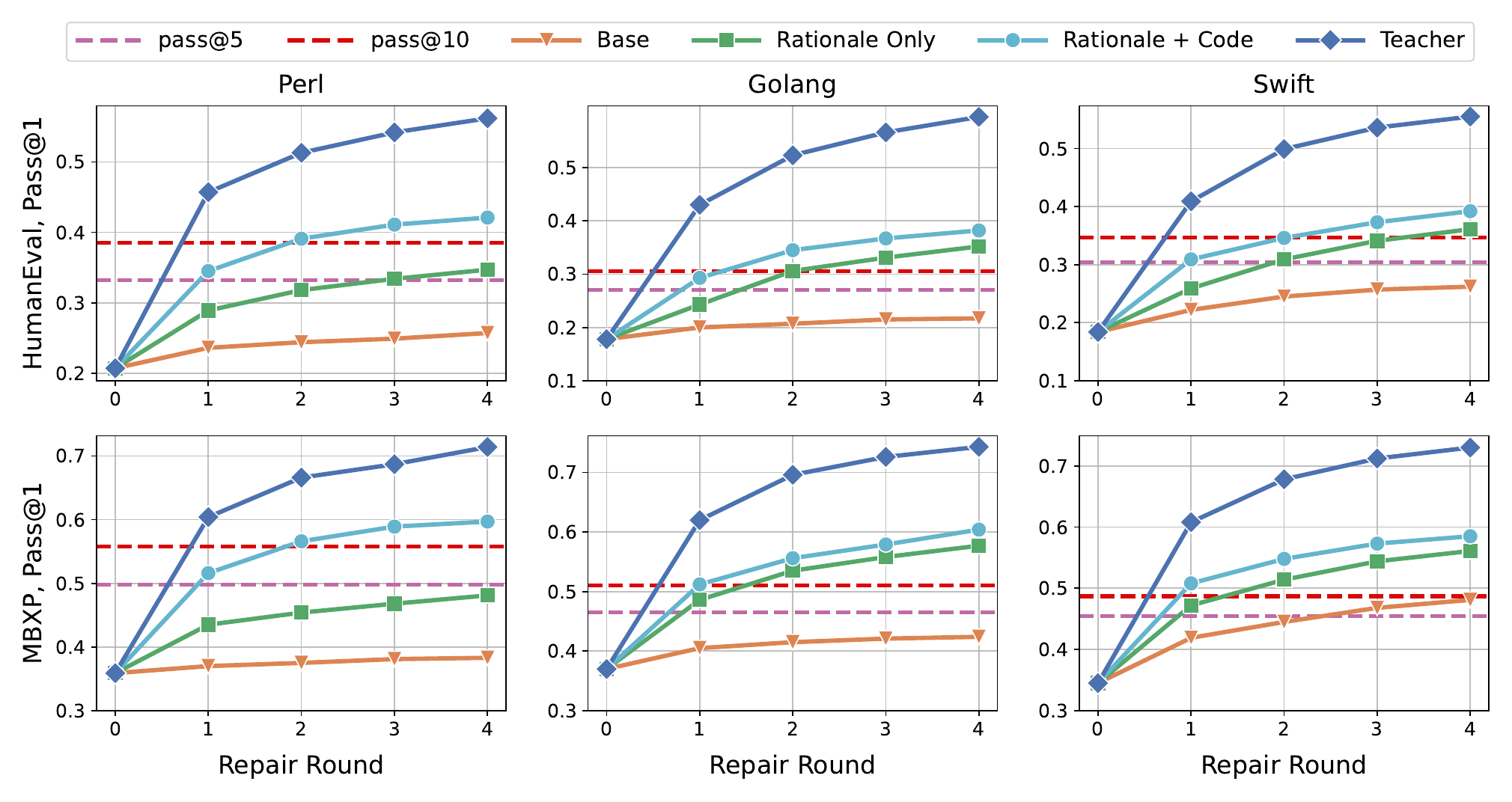}
        \caption{Low-resource languages}
    \end{subfigure}
    \begin{subfigure}[b]{\textwidth}
        \centering
        \includegraphics[width=0.95\textwidth]{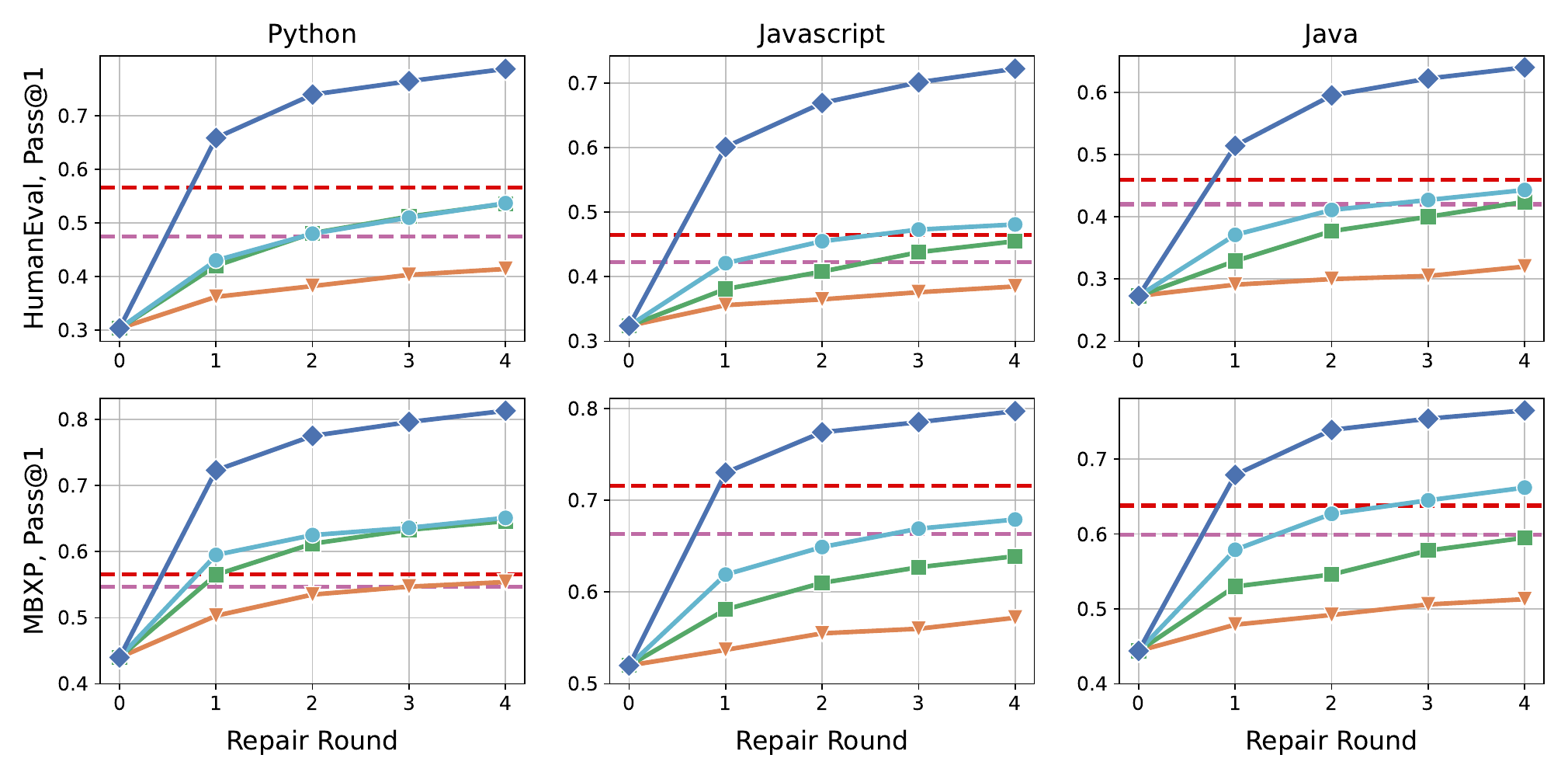}
        \caption{High-resource languages}
    \end{subfigure}
    \caption{Mean pass@1 versus repair round for CodeLlama-7b. Round 0 denotes the initial generation. Rationale-plus-code distillation continues to outperform on all LRPLs. Compared to CodeLlama-7b-Instruct, transferring code completions sees better improvements in HRPLs. One possible explanation for this is that the base CodeLlama-7b has weaker responsiveness to our prompting, due to not being instruction-tuned.}
    \label{fig:plots_clb}
\end{figure}

\newpage
\section{Evaluation on Mistral-7b}
\label{app:evaluation_mist}
\begin{figure}[htbp]
    \centering
    \begin{subfigure}[b]{\textwidth}
        \centering
        \includegraphics[width=0.95\textwidth]{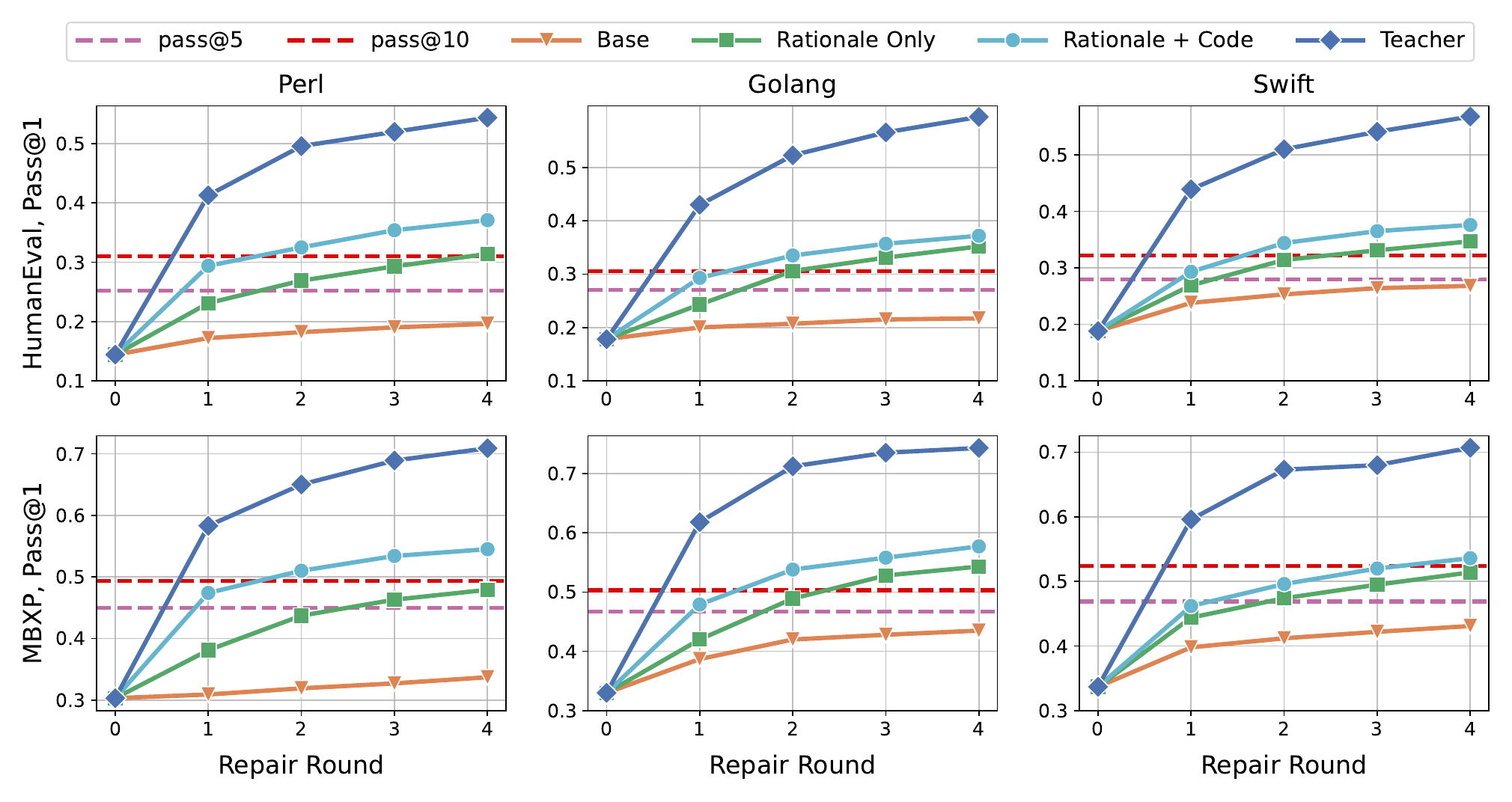}
        \caption{Low-resource languages}
    \end{subfigure}
    \begin{subfigure}[b]{\textwidth}
        \centering
        \includegraphics[width=0.95\textwidth]{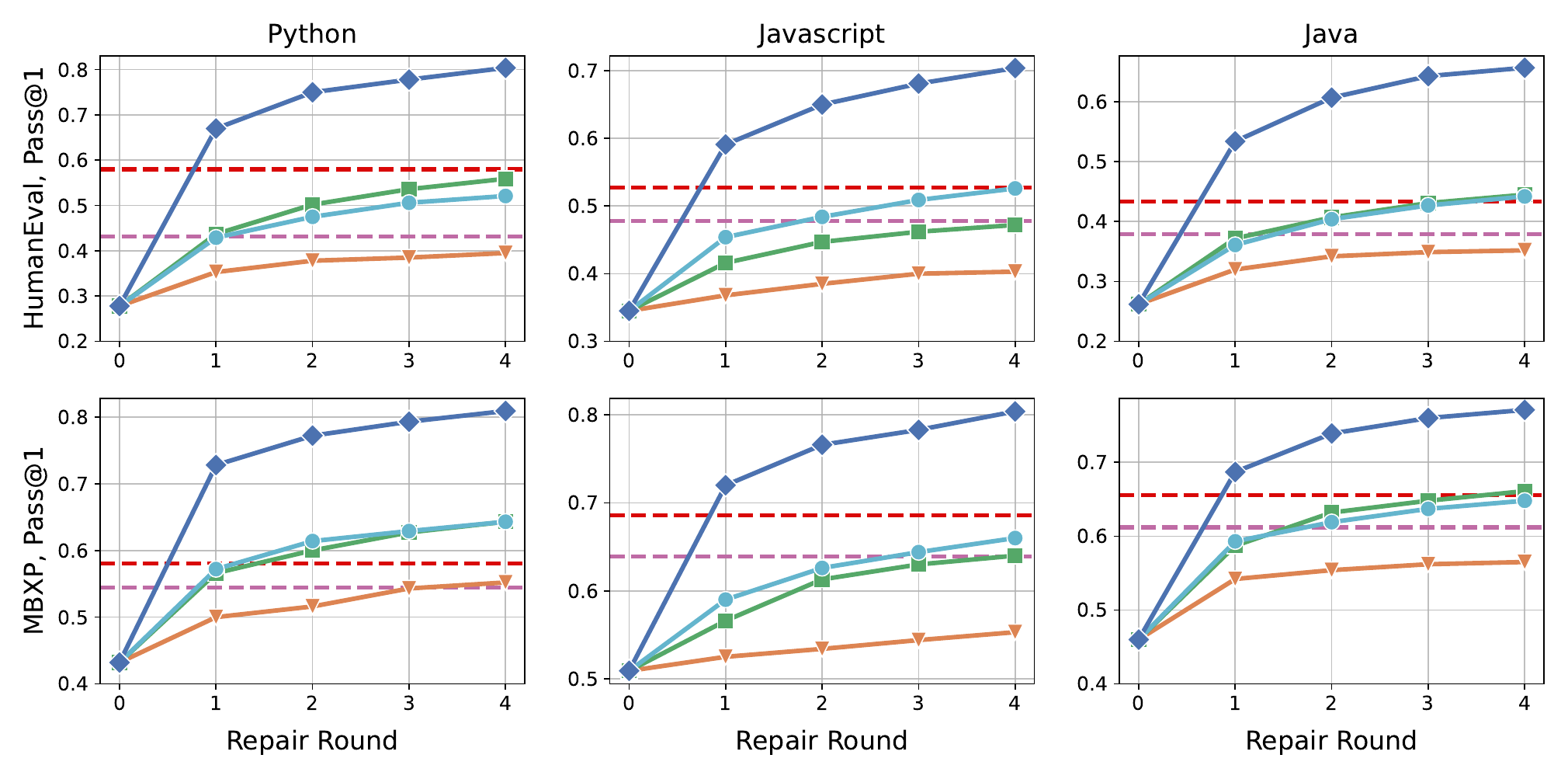}
        \caption{High-resource languages}
    \end{subfigure}
    \caption{Mean pass@1 versus repair round for Mistral-7b. Round 0 denotes the initial generation. Rationale-plus-code distillation continues to outperforms on all LRPLs. Pass@1 improvements between LRPLs and HRPLs trend similarly to CodeLlama-7b-Instruct.}
    \label{fig:plots_mist}
\end{figure}

\newpage
\section{Pass@1 Comparison on CodeLlama-7b}
\begin{table}[htbp]
  \centering
  \resizebox{0.4\textwidth}{!}{
  \begin{tabular}{l|ccc}
    \hline
    \multicolumn{4}{c}{\textbf{HumanEval Pass@1}} \\
    \hline
    \small Language & \small Initial & \small Rationale Only & \small Rationale + Code \\
    \hline
    Perl         & 0.207       & 0.347     & \textbf{0.421} \small \textcolor{mid_green}{$\hspace{0.2em}\uparrow$21.3\%} \\
    Golang       & 0.178       & 0.352     & \textbf{0.372} \small \textcolor{mid_green}{$\hspace{0.2em}\uparrow$5.68\%} \\
    Swift        & 0.184       & 0.361     & \textbf{0.392} \small \textcolor{mid_green}{$\hspace{0.2em}\uparrow$8.58\%}\\
    \hline
    Python       & 0.303       & 0.536     & \textbf{0.537} \small \textcolor{mid_green}{$\hspace{0.2em}\uparrow$1.86\%}\\
    Javascript   & 0.324       & 0.455     & \textbf{0.481} \small \textcolor{mid_green}{$\hspace{0.2em}\uparrow$5.71\%}\\
    Java         & 0.273       & 0.424     & \textbf{0.443} \small \textcolor{mid_green}{$\hspace{0.2em}\uparrow$4.48\%}\\
    \hline
    \multicolumn{4}{c}{\textbf{MBXP Pass@1}} \\
    \hline
    Perl         & 0.359       & 0.481 & \textbf{0.597} \small \textcolor{mid_green}{$\hspace{0.2em}\uparrow$24.1\%} \\
    Golang       & 0.370       & 0.597 & \textbf{0.604} \small \textcolor{mid_green}{$\hspace{0.2em}\uparrow$1.17\%} \\
    Swift        & 0.345       & 0.561 & \textbf{0.585} \small \textcolor{mid_green}{$\hspace{0.2em}\uparrow$4.27\%}\\
    \hline
    Python       & 0.440       & 0.646 & \textbf{0.651} \small \textcolor{mid_green}{$\hspace{0.2em}\uparrow$0.77\%}\\
    Javascript   & 0.520       & 0.639 & \textbf{0.679} \small \textcolor{mid_green}{$\hspace{0.2em}\uparrow$6.25\%}\\
    Java         & 0.444       & 0.595 & \textbf{0.662} \small \textcolor{mid_green}{$\hspace{0.2em}\uparrow$11.2\%}\\
    \hline
  \end{tabular}
  }
  \caption{Mean pass@1 of initial generations vs the two variants of distillation. Rationale-plus-code distillation sees consistent improvements on both LRPLs and HRPLs, likely because CodeLlama-7b has weaker instruction following, so fine-tuning improves its responsiveness.}
  \label{tab:increase_clb}
\end{table}

\section{Pass@1 Comparison on Mistral-7b}
\begin{table}[htbp]
  \centering
  \resizebox{0.4\textwidth}{!}{
  \begin{tabular}{l|ccc}
    \hline
    \multicolumn{4}{c}{\textbf{HumanEval Pass@1}} \\
    \hline
    \small Language & \small Initial & \small Rationale Only & \small Rationale + Code \\
    \hline
    Perl         & 0.144       & 0.314    & \textbf{0.371} \small \textcolor{mid_green}{$\hspace{0.2em}\uparrow$18.1\%} \\
    Golang       & 0.140       & 0.310    & \textbf{0.321} \small \textcolor{mid_green}{$\hspace{0.2em}\uparrow$3.55\%} \\
    Swift        & 0.188       & 0.357     & \textbf{0.366} \small \textcolor{mid_green}{$\hspace{0.2em}\uparrow$2.52\%}\\
    \hline
    Python       & 0.278       & \textbf{0.559}      & 0.520 \small \textcolor{orange}{$\hspace{0.2em}\downarrow$6.97\%}\\
    Javascript   & 0.345       & 0.472     & \textbf{0.526} \small \textcolor{mid_green}{$\hspace{0.2em}\uparrow$11.4\%}\\
    Java         & 0.262       & \textbf{0.445}     & 0.442 \small \textcolor{orange}{$\hspace{0.2em}\downarrow$0.67\%}\\
    \hline
    \multicolumn{4}{c}{\textbf{MBXP Pass@1}} \\
    \hline
    Perl         & 0.303       & 0.479 & \textbf{0.545} \small \textcolor{mid_green}{$\hspace{0.2em}\uparrow$13.7\%} \\
    Golang       & 0.330       & 0.543 & \textbf{0.576} \small \textcolor{mid_green}{$\hspace{0.2em}\uparrow$6.08\%} \\
    Swift        & 0.337       & 0.514 & \textbf{0.536} \small \textcolor{mid_green}{$\hspace{0.2em}\uparrow$4.28\%}\\
    \hline
    Python       & 0.432       & \textbf{0.643} & \textbf{0.643} \small \textcolor{low_orange}{$\hspace{0.2em}\downarrow$0.00\%}\\
    Javascript   & 0.509       & 0.640 & \textbf{0.660} \small \textcolor{mid_green}{$\hspace{0.2em}\uparrow$3.12\%}\\
    Java         & 0.460       & \textbf{0.661} & 0.648 \small \textcolor{low_orange}{$\hspace{0.2em}\downarrow$2.73\%}\\
    \hline
  \end{tabular}
  }
  \caption{Mean pass@1 of initial generations vs the two variants of distillation. A similar pattern as CodeLlama-7b-Instruct is observed, where rationale-plus-code distillation outperforms on LRPLs, but struggles to make consistent improvements on HRPLs.}
  \label{tab:increase_mist}
\end{table}

\section{GPT-4 Judgement Prompt}
\label{app:prompt_judgement}
\begin{figure}[htbp]
  \centering
  \includegraphics[width=0.7\columnwidth]{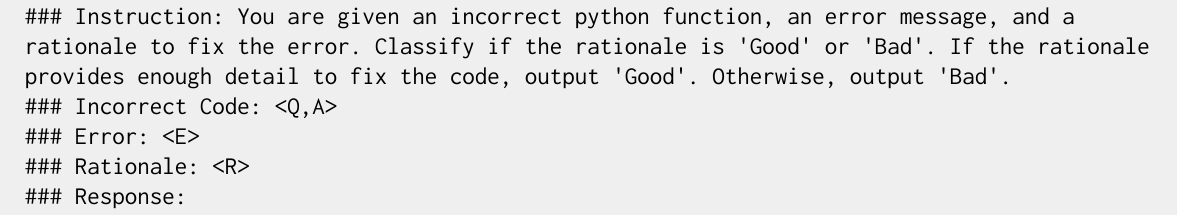}
  \caption{The prompt given to GPT-4 to judge rationale sufficiency. <Q,A> is replaced with the question and previous answer, <E> is replaced with the corresponding error, and <R> is replaced with the repair model's rationale.}
\end{figure}

\newpage
\section{Syntax Errors for CodeLlama-7b}
\label{tab:syntax_clb}

\begin{table}[ht]
  \centering
  \resizebox{0.65\textwidth}{!}{
  \begin{tabular}{c | cllll}
    \hline
    \multicolumn{6}{c}{\textbf{HumanEval Average Syntax Errors}} \\
    \hline
    \multicolumn{1}{c|}{} & \multicolumn{1}{c}{Initial} & \multicolumn{1}{c}{Base} & \multicolumn{1}{c}{Rationale} & \multicolumn{1}{c}{Rationale} & \multicolumn{1}{c}{Teacher} \\
    \multicolumn{1}{c|}{Language} & \multicolumn{1}{c}{Errors} & \multicolumn{1}{c}{Repair} & \multicolumn{1}{c}{Only} & \multicolumn{1}{c}{+ Code} & \multicolumn{1}{c}{Repair}\\
    \hline
    Perl         & 21.2 & 21.0 \small \textcolor{lightergreen}{$\downarrow$0.2} & 20.6 \small \textcolor{lightergreen}{$\downarrow$0.6} & 12.8 \small \textcolor{lightergreen}{$\downarrow$8.4} & \textbf{7.2} \small \textcolor{lightergreen}{$\downarrow$14.0} \\
    Golang       & 39.1 & 72.8 \small \textcolor{lighterred}{$\uparrow$33.7} & 36.6 \small \textcolor{lightergreen}{$\downarrow$2.5} & 30.8 \small \textcolor{lightergreen}{$\downarrow$8.3} & \textbf{22.2} \small \textcolor{lightergreen}{$\downarrow$16.9} \\
    Swift        & 78.1  & 57.2 \small \textcolor{lightergreen}{$\downarrow$20.9} & 47.0 \small \textcolor{lightergreen}{$\downarrow$31.1} & 48.4 \small \textcolor{lightergreen}{$\downarrow$29.7} & \textbf{40.4} \small \textcolor{lightergreen}{$\downarrow$37.7} \\
    \hline
    Python       & 17.1 & 22.4 \small \textcolor{lighterred}{$\uparrow$5.3} & 23.5 \small \textcolor{lighterred}{$\uparrow$6.4} & 12.5 \small \textcolor{lightergreen}{$\downarrow$4.6} & \textbf{7.8} \small \textcolor{lightergreen}{$\downarrow$9.3} \\
    Javascript   & 10.6 & 10.0 \small \textcolor{lightergreen}{$\downarrow$0.6} & 28.6 \small \textcolor{lighterred}{$\uparrow$18.0} & 13.2 \small \textcolor{lighterred}{$\uparrow$2.6} & \textbf{5.4} \small \textcolor{lightergreen}{$\downarrow$5.2} \\
    Java         & 44.7 & 55.0 \small \textcolor{lighterred}{$\uparrow$10.3} & 44.6 \small \textcolor{lightergreen}{$\downarrow$0.1} & 42.4 \small \textcolor{lightergreen}{$\downarrow$2.3} & \textbf{20.4} \small \textcolor{lightergreen}{$\downarrow$24.3}\\
    \hline
    \multicolumn{6}{c}{\textbf{MBXP Average Syntax Errors}} \\
    \hline
    Perl         & 16.4 & 15.0 \small \textcolor{lightergreen}{$\downarrow$1.4} & 15.7 \small \textcolor{lightergreen}{$\downarrow$0.7} & 6.2 \small \textcolor{lightergreen}{$\downarrow$10.2} & \textbf{3.5} \small \textcolor{lightergreen}{$\downarrow$12.9}  \\
    Golang       & 30.7 & 48.2 \small \textcolor{lighterred}{$\uparrow$17.5} & 18.2 \small \textcolor{lightergreen}{$\downarrow$12.5} & 15.2 \small \textcolor{lightergreen}{$\downarrow$15.2} & \textbf{12.6} \small \textcolor{lightergreen}{$\downarrow$18.1} \\
    Swift        & 62.4 & 37.6 \small \textcolor{lightergreen}{$\downarrow$24.8} & 22.6 \small \textcolor{lightergreen}{$\downarrow$39.8} & \textbf{19.0} \small \textcolor{lightergreen}{$\downarrow$43.4} & 19.4 \small \textcolor{lightergreen}{$\downarrow$43.0} \\
    \hline
    Python       & 2.3 & 7.4 \small \textcolor{lighterred}{$\uparrow$5.1} & 5.7 \small \textcolor{lighterred}{$\uparrow$3.4} & 2.5 \small \textcolor{lighterred}{$\uparrow$0.2} & \textbf{1.3} \small \textcolor{lightergreen}{$\downarrow$1.0} \\
    Javascript   & 7.1 & 5.0 \small \textcolor{lightergreen}{$\downarrow$2.1} & 14.6 \small \textcolor{lighterred}{$\uparrow$7.5} & 7.0 \small \textcolor{lightergreen}{$\downarrow$0.1} & \textbf{2.0} \small \textcolor{lightergreen}{$\downarrow$5.1} \\
    Java         & 31.4 & 33.2 \small \textcolor{lighterred}{$\uparrow$1.8} & 24.0 \small \textcolor{lightergreen}{$\downarrow$7.4} & 20.6 \small \textcolor{lightergreen}{$\downarrow$10.8} & \textbf{9.6} \small \textcolor{lightergreen}{$\downarrow$21.8} \\
    \hline
  \end{tabular}
  }
  \caption{Average number of syntax errors after code repair for CodeLlama-7b. We also include a column containing results from the GPT-3.5-Turbo teacher.}
\end{table}

\section{Syntax Errors for Mistral-7b}
\label{tab:syntax_mist}

\begin{table}[ht]
  \centering
  \resizebox{0.65\textwidth}{!}{
  \begin{tabular}{c | cllll}
    \hline
    \multicolumn{6}{c}{\textbf{HumanEval Average Syntax Errors}} \\
    \hline
    \multicolumn{1}{c|}{} & \multicolumn{1}{c}{Initial} & \multicolumn{1}{c}{Base} & \multicolumn{1}{c}{Rationale} & \multicolumn{1}{c}{Rationale} & \multicolumn{1}{c}{Teacher} \\
    \multicolumn{1}{c|}{Language} & \multicolumn{1}{c}{Errors} & \multicolumn{1}{c}{Repair} & \multicolumn{1}{c}{Only} & \multicolumn{1}{c}{+ Code} & \multicolumn{1}{c}{Repair}\\
    \hline
    Perl         & 26.4 & 30.4 \small \textcolor{lighterred}{$\uparrow$4.0} & 24.0 \small \textcolor{lightergreen}{$\downarrow$2.4} & 11.0 \small \textcolor{lightergreen}{$\downarrow$15.4} & \textbf{9.4} \small \textcolor{lightergreen}{$\downarrow$17.0} \\
    Golang       & 55.9 & 72.4 \small \textcolor{lighterred}{$\uparrow$16.5} & 48.2 \small \textcolor{lightergreen}{$\downarrow$7.7} & 31.0 \small \textcolor{lightergreen}{$\downarrow$24.9} & \textbf{25.2} \small \textcolor{lightergreen}{$\downarrow$30.7} \\
    Swift        & 62.0  & 60.0 \small \textcolor{lightergreen}{$\downarrow$2.0} & 54.0 \small \textcolor{lightergreen}{$\downarrow$8.0} & 55.4 \small \textcolor{lightergreen}{$\downarrow$6.6} & \textbf{39.8} \small \textcolor{lightergreen}{$\downarrow$22.2} \\
    \hline
    Python       & 14.5 & 17.2 \small \textcolor{lighterred}{$\uparrow$2.7} & 12.4 \small \textcolor{lightergreen}{$\downarrow$2.1} & 13.5 \small \textcolor{lightergreen}{$\downarrow$1.0} & \textbf{8.0} \small \textcolor{lightergreen}{$\downarrow$6.5} \\
    Javascript   & 6.7 & \textbf{7.4} \small \textcolor{lighterred}{$\uparrow$0.7} & 16.6 \small \textcolor{lighterred}{$\uparrow$9.9} & 7.8 \small \textcolor{lighterred}{$\uparrow$1.1} & 7.8 \small \textcolor{lighterred}{$\uparrow$1.1} \\
    Java         & 41.4 & 42.2 \small \textcolor{lighterred}{$\uparrow$0.8} & 36.2 \small \textcolor{lightergreen}{$\downarrow$5.2} & 31.4 \small \textcolor{lightergreen}{$\downarrow$10.0} & \textbf{19.2} \small \textcolor{lightergreen}{$\downarrow$22.2}\\
    \hline
    \multicolumn{6}{c}{\textbf{MBXP Average Syntax Errors}} \\
    \hline
    Perl         & 26.3 & 25.2 \small \textcolor{lightergreen}{$\downarrow$1.1} & 24.0 \small \textcolor{lightergreen}{$\downarrow$2.3} & 6.0 \small \textcolor{lightergreen}{$\downarrow$20.3} & \textbf{4.2} \small \textcolor{lightergreen}{$\downarrow$22.1}  \\
    Golang       & 43.6 & 40.0 \small \textcolor{lightergreen}{$\downarrow$3.6} & 27.2 \small \textcolor{lightergreen}{$\downarrow$16.4} & 13.8 \small \textcolor{lightergreen}{$\downarrow$29.8} & \textbf{13.2} \small \textcolor{lightergreen}{$\downarrow$30.4} \\
    Swift        & 49.3 & 36.6 \small \textcolor{lightergreen}{$\downarrow$12.7} & 32.6 \small \textcolor{lightergreen}{$\downarrow$16.7} & 25.6 \small \textcolor{lightergreen}{$\downarrow$23.7} & \textbf{21.6} \small \textcolor{lightergreen}{$\downarrow$27.7} \\
    \hline
    Python       & 0.9 & 3.6 \small \textcolor{lighterred}{$\uparrow$2.7} & 4.2 \small \textcolor{lighterred}{$\uparrow$3.3} & 3.2 \small \textcolor{lighterred}{$\uparrow$2.3} & \textbf{2.2} \small \textcolor{lighterred}{$\uparrow$1.3} \\
    Javascript   & 7.8 & 7.6 \small \textcolor{lightergreen}{$\downarrow$0.2} & 10.4 \small \textcolor{lighterred}{$\uparrow$2.6} & 7.4 \small \textcolor{lightergreen}{$\downarrow$0.4} & \textbf{3.6} \small \textcolor{lightergreen}{$\downarrow$4.2} \\
    Java         & 29.9 & 15.0 \small \textcolor{lightergreen}{$\downarrow$14.9} & 11.2 \small \textcolor{lightergreen}{$\downarrow$18.7} & 14.0 \small \textcolor{lightergreen}{$\downarrow$15.9} & \textbf{7.2} \small \textcolor{lightergreen}{$\downarrow$22.7} \\
    \hline
  \end{tabular}
  }
  \caption{Average number of syntax errors after code repair for Mistral-7b. We also include a column containing results from the GPT-3.5-Turbo teacher.}
\end{table}

\newpage
\section{Distillation vs Baselines Examples}
\label{app:actual_examples}
Our work hypothesizes that base models struggle to convert suggestions from the rationale into specific code corrections. We provide examples of this for each language. In particular, we highlight the differences between models distilled only rationales vs models distilled both rationales and code completions. In each example, the rationale-only model is provided a stellar rationale but generates incorrect code. Meanwhile, the rationale-plus-code model produces a weaker rationale but generates correct code. These examples support that the correlation between rationale quality and code correctness is weaker than previously perceived, and that distilling both rationales and code teaches models to respond better to feedback.

\begin{figure}[ht]
  \centering
  \includegraphics[width=\columnwidth]{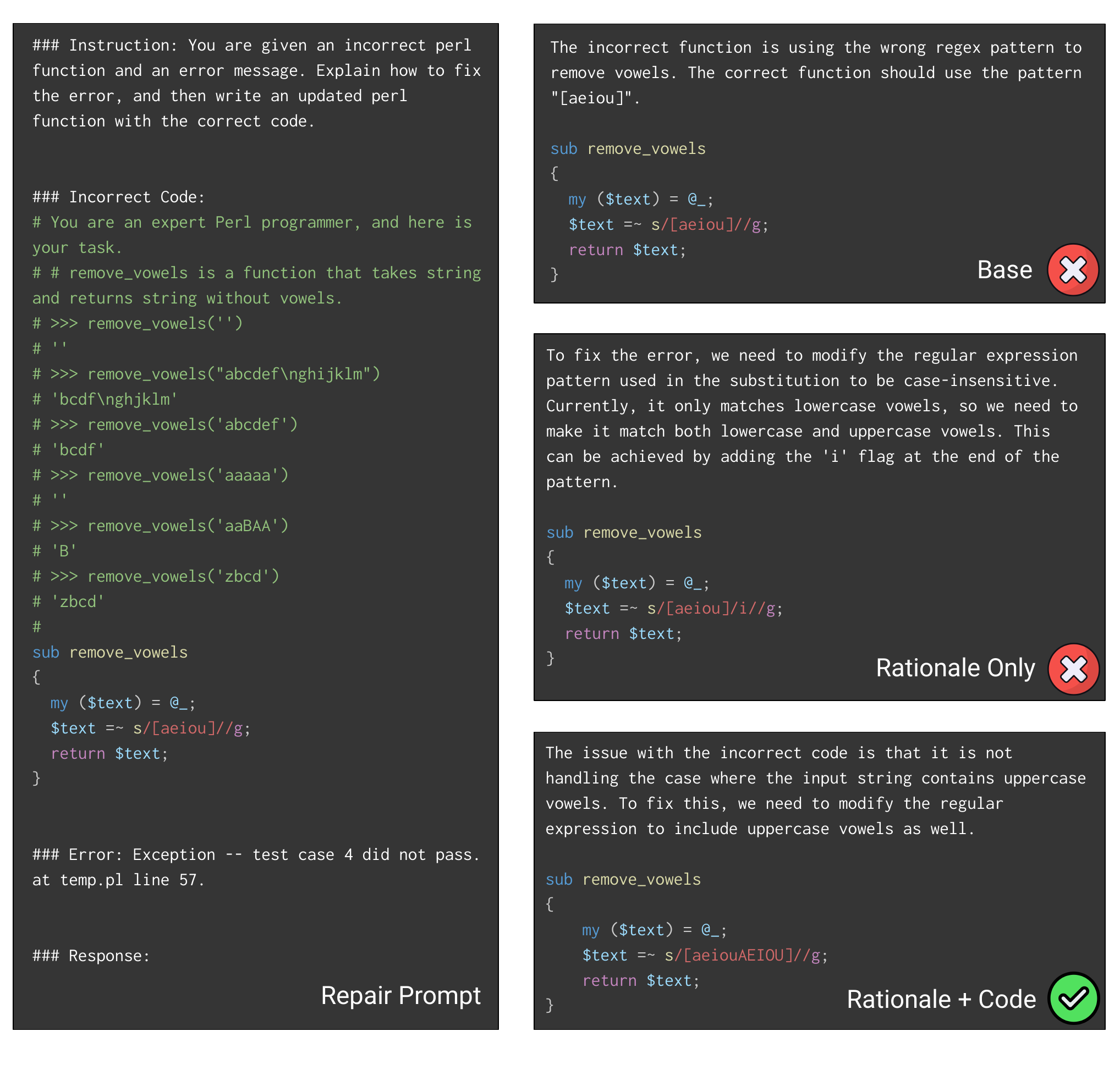}
  \caption{Perl example from HumanEval question 51. The initial code is wrong because it does not remove uppercase vowels. From the base model, we see a weak rationale that fails to diagnose the uppercase issue. From the rationale-only model, we see a stellar rationale that proposes using the "i" regex flag for case insensitivity. However, the generated code incorrectly modifies the regex to "/[aeiou]/i//g" instead of "/[aeiou]//gi", displaying a lack of knowledge. From the rationale-plus-code model, we see a good rationale that suggests adding uppercase letters to the regex, followed by correct code modifications.}
  \label{fig:example_perl}
\end{figure}

\begin{figure}[ht]
  \centering
  \includegraphics[width=\columnwidth]{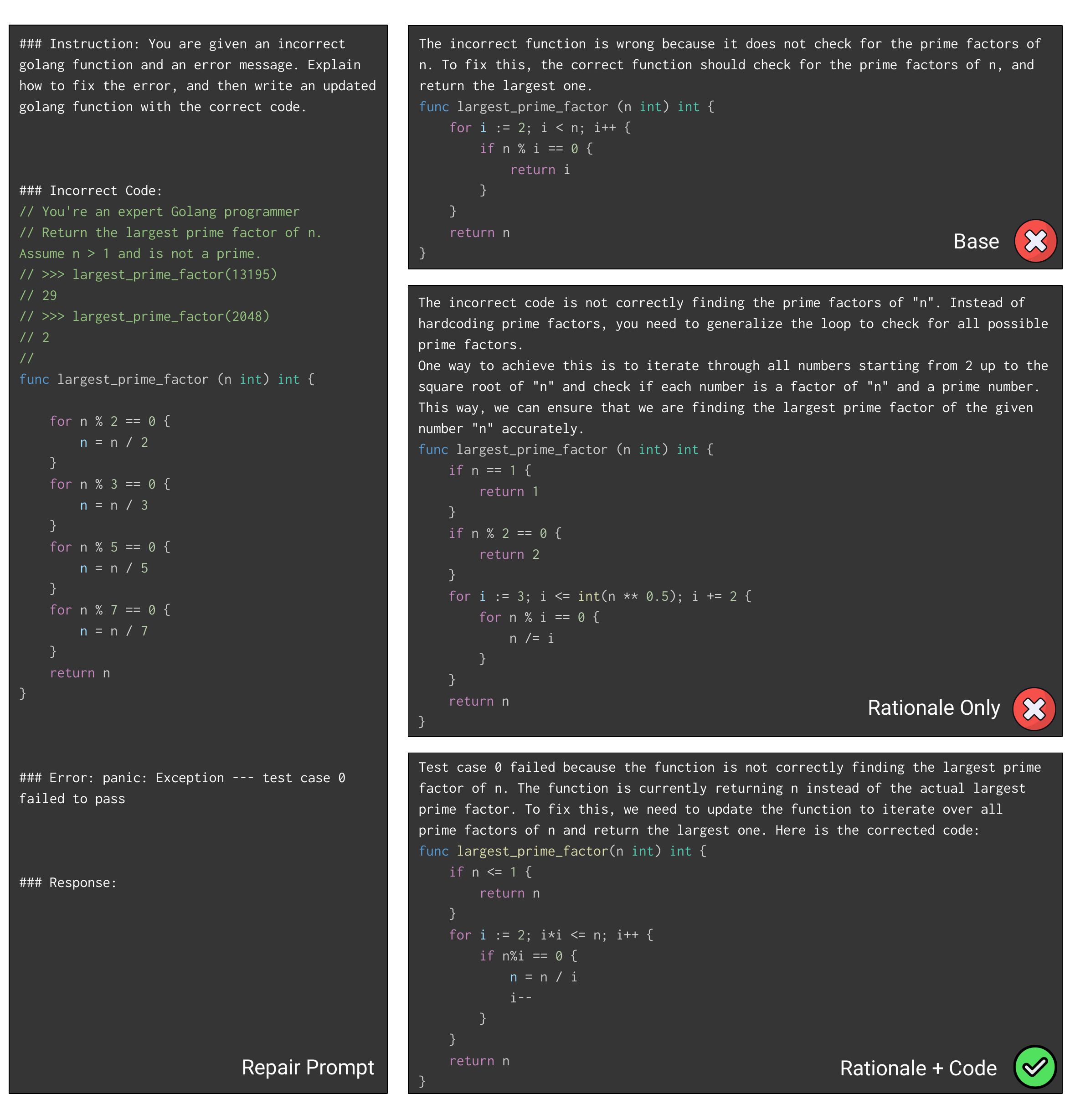}
  \caption{Golang example from HumanEval question 59. The initial code is wrong because it hardcodes prime factors. From the base model, we see a weak rationale that is unable to diagnose the hardcoding issue. From the rationale-only model, we see a stellar rationale that suggests iterating over all numbers from 2 to $\sqrt{\text{n}}$. However, the generated code results in a compilation error due to using "n $**$ 0.5" to obtain the square root (which does not work in golang), displaying a lack of knowledge. From the rationale-plus-code model, we see a good rationale that provides a plan of iterating over prime factors, followed by correct code modifications.}
  \label{fig:example_golang}
\end{figure}

\begin{figure}[ht]
  \centering
  \includegraphics[width=\columnwidth]{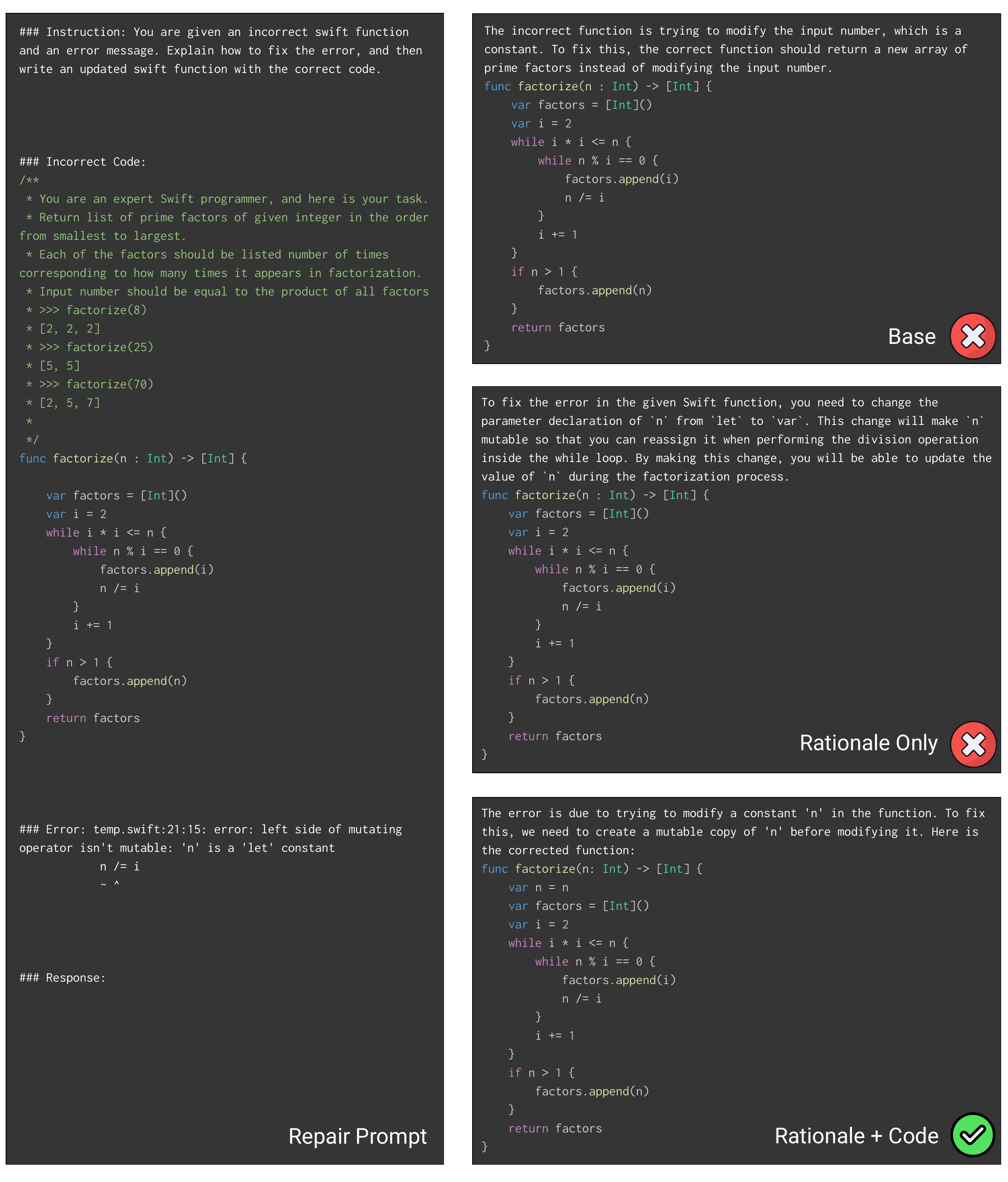}
  \caption{Swift example from HumanEval question 25. The initial code is wrong because it tries to modify the immutable input variable n. For swift, input parameters are immutable by default. From the base model, we see a weak rationale that does not specify how to fix the error. From the rationale-only model, we see a stellar rationale that explicitly mentions using the "var" keyword. However, the generated code remains the same because the base model is unable to make the necessary code modifications, displaying a lack of responsiveness. From the rationale-plus-code model, we see a good rationale that proposes making a mutable copy, followed by correct code modifications through the additional line "var n = n".}
  \label{fig:example_swift}
\end{figure}

\begin{figure}[ht]
  \centering
  \includegraphics[width=\columnwidth]{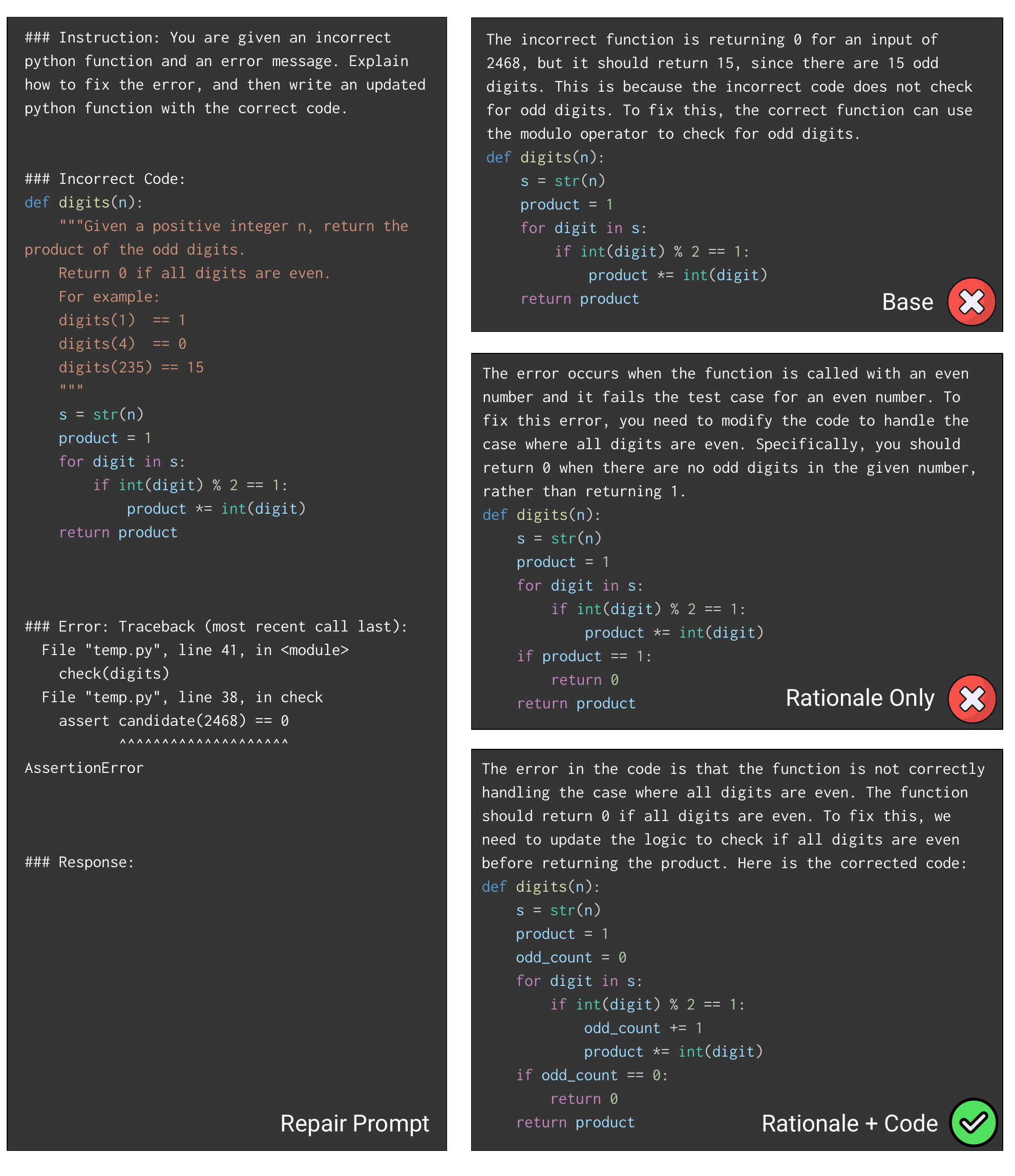}
  \caption{Python example from HumanEval question 131. The initial code is wrong because it returns 1 if there are no odd digits, instead of 0. From the base model, we see a weak rationale that states wrong information. From the rationale-only model, we see a stellar rationale that addresses the edge case and suggests returning 0 if there are no odd digits. However, the generated code implements this incorrectly, displaying weak responsiveness. From the rationale-plus-code model, we see a good rationale that suggests checking if all digits are even, followed by correct code modifications via the additional variable "odd\_count".}
  \label{fig:example_python}
\end{figure}

\begin{figure}[ht]
  \centering
  \includegraphics[width=\columnwidth]{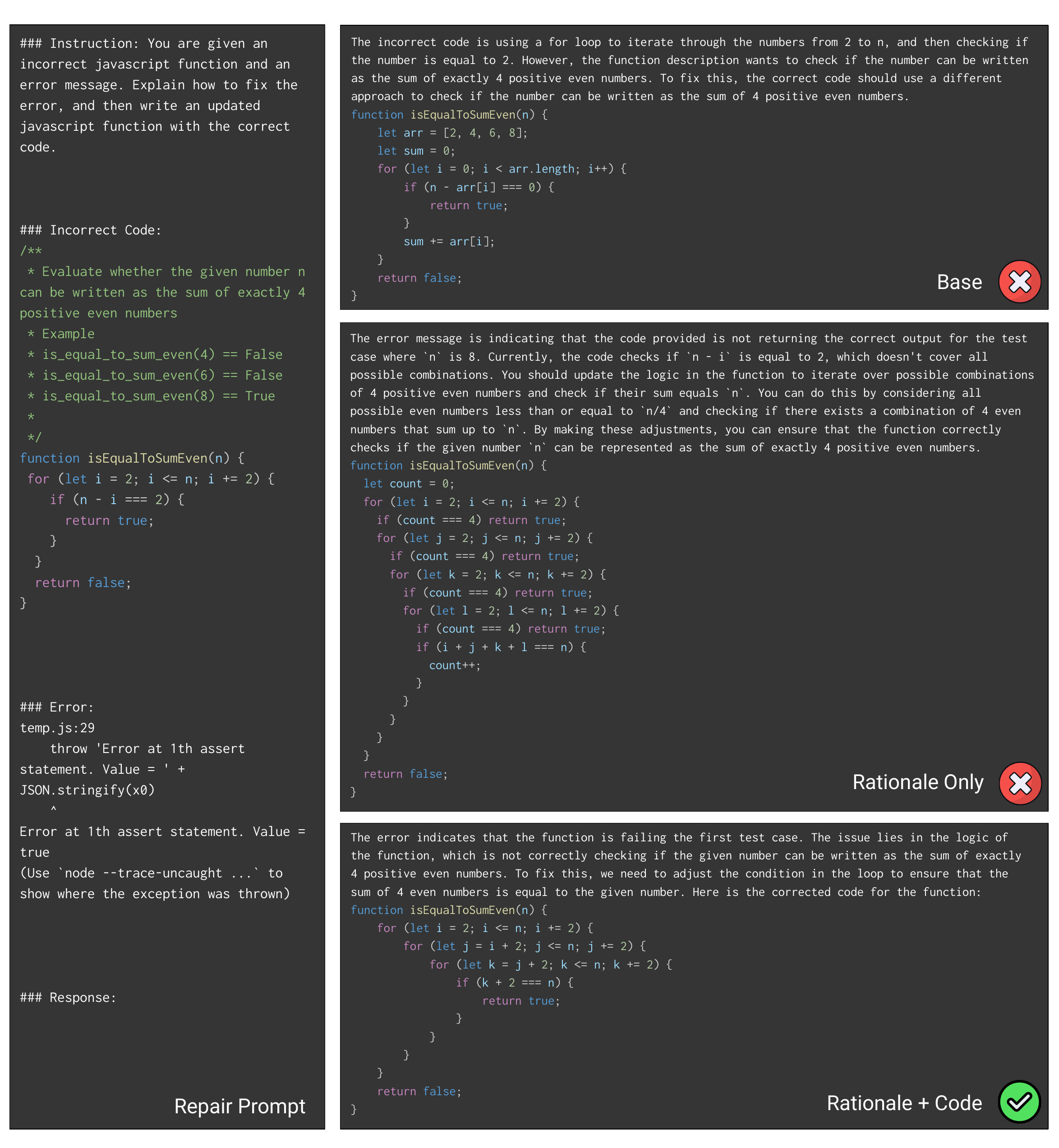}
  \caption{Javascript example from HumanEval question 138. The initial code is wrong because it checks if n is a sum of 2 even numbers, instead of 4. From the base model, we see a decent rationale that lacks a highly detailed plan. From the rationale-only model, we see a stellar rationale suggesting a highly detailed plan. However, the generated code implements this incorrectly by misusing the "count" variable, displaying weak responsiveness. From the rationale-plus-code model, we see a decent rationale that lacks a highly detailed plan, but correct code modifications anyways, displaying strong responsiveness.}
  \label{fig:example_javascript}
\end{figure}

\begin{figure}[ht]
  \centering
  \includegraphics[width=\columnwidth]{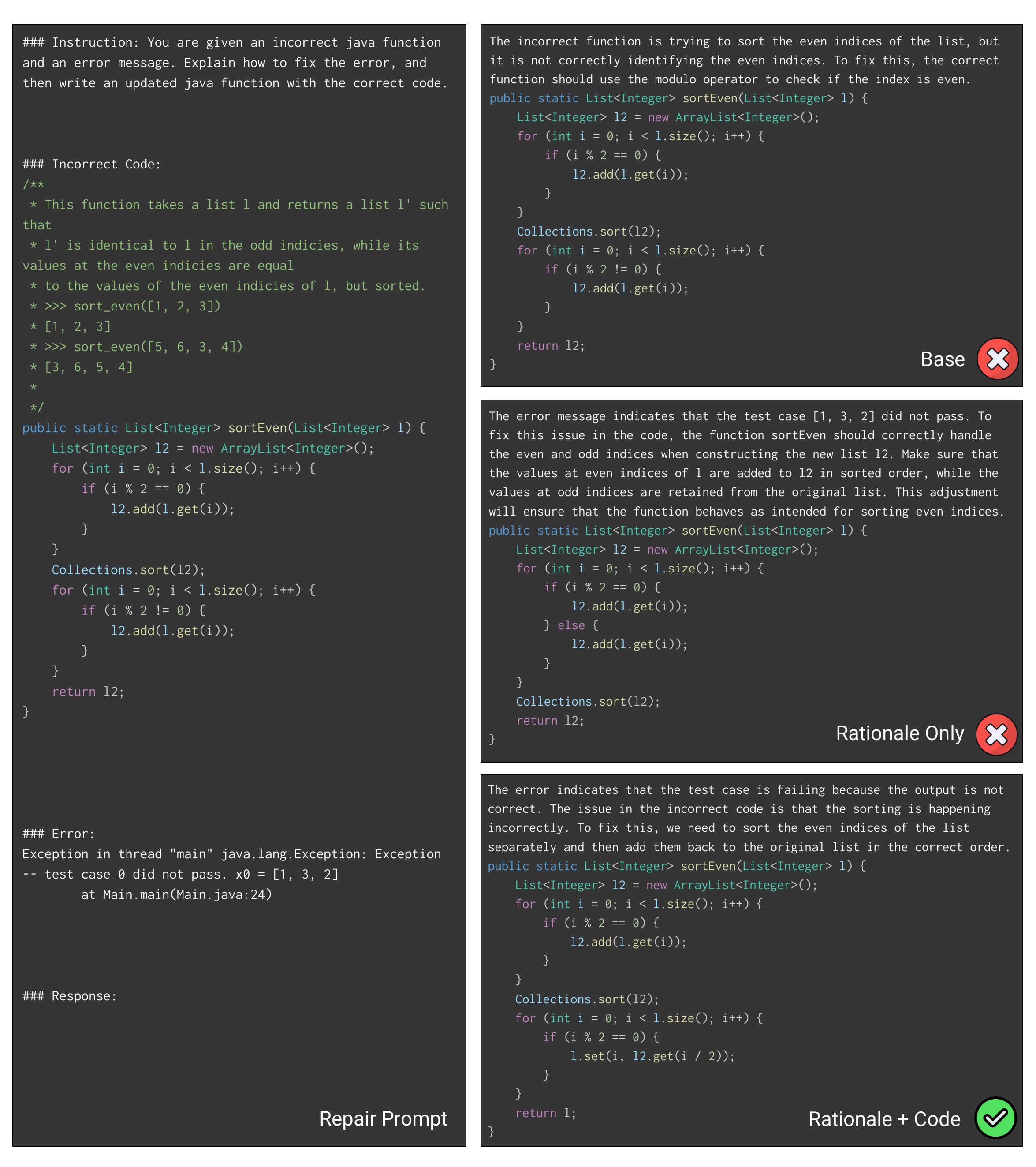}
  \caption{Java example from HumanEval question 37. The initial code is wrong because it first appends even indices and then appends odd indices, instead of interleaving them. From the base model, we see a weak rationale and no code modifications. From the rationale-only model, we see a stellar rationale suggesting to interleave odd/even indices. However, the generated code incorrectly implements the sorting of even indices by sorting the entire list at the end, displaying weak responsiveness. From the rationale-plus-code model, we see a good rationale suggesting to interleave odd/even indices, followed by correct code modifications.}
  \label{fig:example_java}
\end{figure}

\end{document}